\documentclass[lettersize,journal]{IEEEtran}
\usepackage{amsmath,amssymb,amsfonts} 
\usepackage{algorithmic}
\usepackage{algorithm}
\usepackage{array}
\usepackage[caption=false,font=normalsize,labelfont=sf,textfont=sf]{subfig}
\usepackage{textcomp}
\usepackage{stfloats}
\usepackage{verbatim}
\usepackage{graphicx}
\newcommand{\xing}[1]{\textcolor[rgb]{1.0,0.0,0.0}{\textbf{#1}}}

\usepackage{graphics} 
\usepackage{bm}
\usepackage{multirow}
\usepackage{booktabs}
\usepackage[table,xcdraw]{xcolor}

\usepackage{my_predifined_commands}

\usepackage{url}
\usepackage{hyperref}
\hypersetup{colorlinks,urlcolor=red}

\usepackage[backend=bibtex,natbib=true,style=numeric-comp,sorting=none,giveninits=true,maxbibnames=99,url=false,doi=false]{biblatex}
\begin{document}

\title{ \LARGE 
Continuous-Time Fixed-Lag Smoothing for \\ LiDAR-Inertial-Camera SLAM
}

\author{Jiajun Lv$^{1}$, Xiaolei Lang$^{1}$, Jinhong Xu$^{1}$, Mengmeng Wang$^{1}$, Yong Liu$^{1,2*}$, Xingxing Zuo$^{3,*}$ 
\thanks{$^1$ The authors are with the State Key Laboratory of Industrial Control Technology, Zhejiang University, Hangzhou, China.}
\thanks{$^2$ The author is also with the Huzhou Institute of Zhejiang University.}
\thanks{$^3$ The author is with the School of Computation, Information and Technology, Technical University of Munich, Munich, Germany.}
\thanks{$^*$ Yong Liu and Xingxing Zuo are the corresponding authors (Email: {\tt\small yongliu@iipc.zju.edu.cn; xingxing.zuo@tum.de}).}
}

\markboth{IEEE/ASME Transactions on Mechatronics,~Vol.~xx, No.~x, July~2022}%
{Lv \MakeLowercase{\textit{et al.}}: Continuous-Time Fixed-Lag Smoothing}


\maketitle

\setlength{\textfloatsep}{2.0pt} 

\begin{abstract}
Localization and mapping with heterogeneous multi-sensor fusion have been prevalent in recent years. To adequately fuse multi-modal sensor measurements received at different time instants and different frequencies, we estimate the continuous-time trajectory by fixed-lag smoothing within a factor-graph optimization framework. With the continuous-time formulation, we can query poses at any time instants corresponding to the sensor measurements. To bound the computation complexity of the continuous-time fixed-lag smoother, we maintain temporal and keyframe sliding windows with constant size, and probabilistically marginalize out control points of the trajectory and other states, which allows preserving prior information for future sliding-window optimization. 
Based on continuous-time fixed-lag smoothing, we design tightly-coupled multi-modal SLAM algorithms with a variety of sensor combinations, like the LiDAR-inertial and LiDAR-inertial-camera SLAM  systems, in which online timeoffset calibration is also naturally supported. More importantly, benefiting from the marginalization and our derived analytical Jacobians for optimization, the proposed continuous-time SLAM systems can achieve real-time performance regardless of the high complexity of continuous-time formulation.
The proposed multi-modal SLAM systems have been widely evaluated on three public datasets and self-collect datasets. The results demonstrate that the proposed continuous-time SLAM systems can achieve high-accuracy pose estimations and outperform existing state-of-the-art methods. To benefit the research community, we will open source our code at ~\url{https://github.com/APRIL-ZJU/clic}.
\end{abstract}


\section{Introduction}
\IEEEPARstart{S}{imultaneous} \textbf{L}ocalization \textbf{A}nd \textbf{M}apping (\textbf{SLAM}) are fundamental for mobile robots to navigate autonomously in various applications. Especially, in scenarios where external signals are unavailable, e.g., the GPS-denied environment, localization and mapping with only onboard sensors can be a practical solution. 
Plenty of sensors have been used for SLAM purposes, including the proprioceptive sensors and exteroceptive sensors. The common proprioceptive sensors measure the state of robot, e.g., wheel encoders, \textbf{I}nertial \textbf{M}easurement \textbf{U}nit (\textbf{IMU}), and magnetometer, etc. While the exteroceptive sensors perceive the surrounding environment information, for example, radar, sonar, LiDAR, camera, barometer, altimeter, etc. 
Among all of the sensors, camera, IMU and LiDAR are three of the most ubiquitous sensors leveraged in SLAM algorithms~\cite{murORB2,zhang2014loam,qin2018vins,zuo2019lic,zuo2020multimodal,shan2021lvi}. 

Recently, \textbf{L}iDAR-\textbf{I}nertial-\textbf{C}amera (\textbf{LIC}) based localization and mapping systems~\cite{zuo2019lic, shan2021lvi, lin2021r3live} have attracted significant attention, due to their versatility, high accuracy, and robustness. By combining the three sensor modalities, LIC systems have more applicable scenarios than using only the component sensors.
Besides, since a single LiDAR only has a limited field of view (FOV), multiple LiDARs fusion~\cite{LOCUS} is a rational choice for highly-accurate localization and efficient mapping.

Multi-modal sensor measurements are assigned with timestamps by individual sensors, and timeoffsets generally exist among different sensors~\cite{li2014online,zuo2020lic} without hardware synchronization. Even the timeoffsets between sensors can be removed, measurements from various sensor modalities are usually received at different rates and different time instants. In order to fuse the heterogeneous sensors, accurate alignment of the asynchronous sensors is the prerequisite. 
Without special hardware support for synchronization, timeoffset can also be online calibrated in estimators, such as online timeoffset estimation in \textbf{V}isual-\textbf{I}nertial(\textbf{VI}) system~\cite{li2014online, qin2018online, yang2020online, lang2022ctrl} and LIC system~\cite{zuo2019lic}. 
Further efforts are required to deal with different sensor frequencies and sampling time instants. For example, sensors like IMU and LiDAR consecutively provide abundant measurements at high frequencies, and it is challenging to accurately align thousands of points in LiDAR scans to the asynchronous IMU measurements. Some approaches~\cite{zuo2020lic, shan2021lvi, geneva2018lips} linearly interpolate the integrated discrete-time IMU poses at the sampling time instants of LiDAR points, in order to compensate for motion distortion in LiDAR scans. 

Another alternative way is to parameterize the trajectory in a continuous-time representation~\cite{sommer2020efficient,lowe2018complementary,lv2021clins}, which allows pose querying at any time instants without interpolations.
Formulating the trajectory in continuous-time with B-spline gains popularity in calibration tasks~\cite{rehder2016extending,lv2020targetless}, visual odometry~\cite{mueggler2018continuous}, as well as LiDAR odometry~\cite{lv2021clins, park2021elasticity}, due to its versatility and convenience in aligning asynchronous sensor measurements. 
However, there are two main challenges preventing continuous-time trajectory from being widely deployed: 
\textit{(i) Real-time performance:} In conventional discrete-time state estimation, the pose in the residual is just the state to be estimated. However, in continuous-time state estimation, the pose is computed from multiple control points on the Lie group, and the state to be estimated in the residual switches to multiple control points (depend on the B-spline order). This fact not only increases the computation load but also generally increases the complexity of the Jacobian computation. Existing continuous-time odometry methods~\cite{lv2021clins} rarely achieve real-time performance. 
\textit{(ii) Fixed-lag smoothing:} Discrete-time methods typically estimate states by fixed-lag smoothing~\cite{dong2011motion,leutenegger2015keyframe,qin2018vins} and preserve information of the old measurements/states with marginalization, however, continuous-time methods like VIO~\cite{yang2021asynchronous} or LIO~\cite{park2021elasticity} rarely consider how to preserve information of old measurements/states. In fact, some of informative measurements/states are directly discarded without leaving prior information for the estimation of remaining states. There is little work about marginalization for continuous-time trajectory optimization, probably because of the high complexity of optimizing continuous-time trajectory, and the sliding strategy and marginalization strategy are significantly different from discrete-time methods.

In general, although B-spline based continuous-time trajectory optimization methods have been studied in many existing research works~\cite{yang2021asynchronous, park2021elasticity, lv2021clins}, the continuous-time fixed-lag smoothing within a sliding window and with probabilistic state marginalization is rarely investigated in existing literature.
To the best of our knowledge, this paper is among the \textit{first} to utilize continuous-time fixed-lag smoothing with probabilistic marginalization for multi-sensor fusion, and even better, we can achieve real-time performance.  
The contributions can be summarized as follows:
\begin{itemize}
    \item We adopt continuous-time fixed-lag smoothing method for multi-sensor fusion in a factor-graph optimization framework. Specifically, we estimate B-spline based continuous-time trajectory within a constant size of sliding window by fusing asynchronous heterogeneous sensor measurements published at various frequencies and different time instants.
    To attain a bounded computation complexity, old control points of the continuous-time trajectory are strategically and probabilistically marginalized out of the sliding window. 
    \item Powered by continuous-time fixed-lag smoothing, we design some \textbf{L}iDAR-\textbf{I}nertial (LI) SLAM and LIC SLAM systems at a variety of sensor combinations (even with multiple LiDARs),  and derive the analytical Jacobians for efficient factor-graph optimization. Benefiting from easy accessibility of continuous-time trajectory derivatives, timeoffsets between different sensors can be online calibrated. With careful and strategic implementation, our proposed continuous-time LI SLAM and LIC SLAM systems can achieve real-time performance.
    \item The proposed continuous-time LI SLAM and LIC SLAM systems are extensively evaluated on three publicly available datasets and self-collected datasets. The experimental results show that the proposed LI system has competitive accuracy and the proposed LIC system outperforms several state-of-the-art methods and works well in degenerated sequences.
\end{itemize}
The remainder of the paper is organized as follows: Sec.~\ref{sec:related_works} reviews the relevant literature. Then with the continuous-time trajectory preliminary provided in Sec.~\ref{sec:traj}, we present continuous-time fixed-lag smoothing method leveraged in Sec.~\ref{sec:fixed-lag-smoothing}. In Sec.~\ref{sec:method}, we further detail multi-sensor fusion SLAM systems enabled by continuous-time fixed-lag smoothing with different sensor configurations, such as LiDAR-inertial systems and LiDAR-inertial-camera systems. Sec.~\ref{sec:experiment} demonstrates the performance of different sensor combinations and discloses the runtime of the different systems. Finally, Sec.~\ref{sec:conclusion} concludes the paper and discusses future work.

\section{Related Works} \label{sec:related_works}

\begin{figure}[t]
    \centering
    \includegraphics[width=1\linewidth]{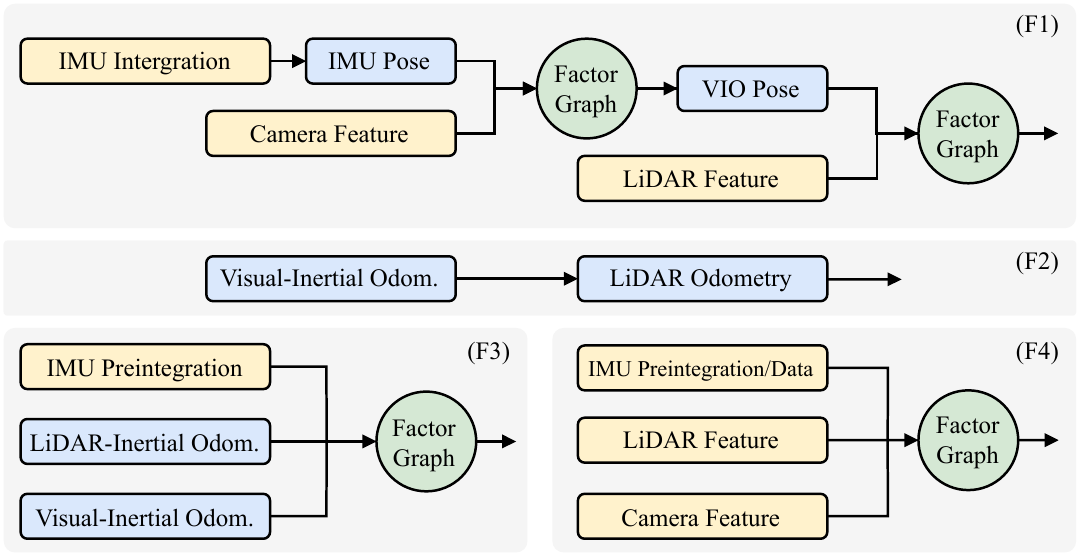}
    \vspace{-1em}
    \captionsetup{font={small}}
    \caption{Four different pipelines of LIC SLAM systems using factor-graph optimization framework. F1, F2, F3 are loosely-coupled methods, and F4 is a tightly-coupled method.}
    \label{fig:four_factor_graph}
    \vspace{-2em}
\end{figure}

\begin{table}[t]
\centering
\caption{Related works of LIC fusion.}
\label{tab:lic_related_works}
\resizebox{\linewidth}{!}{
\begin{tabular}{@{}llcccc@{}}
\toprule
Paper             & Year       & IMU$^{1}$    & Camera$^{2}$ & LiDAR$^{3}$  & Method$^{4}$ \\ \midrule
V-LOAM~\cite{zhang2018laser}            & 2018-JFR   & I1     & C1, C3 & L1     & F1        \\ \midrule
Wang.~\cite{wang2019robust}             & 2019-IROS  & I2     & C1     & L1     & F2        \\ \midrule
Khattak.~\cite{khattak2020complementary}& 2019-ICUAS & \multicolumn{2}{c}{ROVIO~\cite{bloesch2017iterated}}        & L1     &F2         \\ \midrule
Lowe.~\cite{lowe2018complementary}      & 2018-RAL   & I3     & C1, C4 & L5     & F4        \\ \midrule
LIC-Fusion~\cite{zuo2019lic}            & 2019-IROS  & I1     & C1     & L1     & MSCKF     \\ \midrule
LIC-Fusion2.0~\cite{zuo2020lic}         & 2020-IROS  & I1     & C1     & L1, L2 & MSCKF     \\ \midrule
LVI-SAM~\cite{shan2021lvi}              & 2021-ICRA  & I2     & C1, C3 & L1     & F3        \\ \midrule
VILENS~\cite{wisth2021unified}          & 2021-RAL   & I2     & C1, C3 & L2     & F4        \\ \midrule
VILENS~\cite{wisth2021vilens}           & 2021-Arxiv & I2     & C1, C3 & L2, L3 & F4        \\ \midrule
R2live~\cite{lin2021r}                  & 2021-RAL   & I1, I2 & C1     & L1     & ESIKF, F4 \\ \midrule
R3live~\cite{lin2021r3live}             & 2021-Arxiv & I1     & C2, C3 & L1     & ESIKF     \\ \midrule
Super Odom.~\cite{zhao2021super}     & 2021-IROS  & I2     & C1, C3 & L4     & F3        \\ \midrule
Lvio-Fusion~\cite{jia2021lvio}          & 2021-IROS  & I2     & C1     & L1     & F4        \\ \midrule
\multicolumn{6}{l}{$^{1}$ I1=Integration; I2=Preintegration; I3=Raw measurements.} \\
\multicolumn{6}{l}{$^{2}$ C1=Indirect; C2=Direct; C3=Depth From LiDAR; C4=Depth From Surfel.}   \\
\multicolumn{6}{l}{$^{3}$ L1=LOAM Feature; L2=Tracked Plane/Line; L3=PCA based Feature;}  \\
\multicolumn{6}{l}{$\,\,\,$ L4=PCA based Feature; L5=Surfel.} \\
\multicolumn{6}{l}{$^{4}$ See Fig.~\ref{fig:four_factor_graph} for details.} \\ \bottomrule
\end{tabular}
}
\end{table}

There is a rich body of literature on multi-sensor fusion for localization and mapping.
Instead of providing a comprehensive literature review of SLAM with multi-sensor fusion, we extensively review the existing LiDAR-Inertial-Camera systems, multi-LiDAR systems and continuous-time trajectory based SLAM systems, which are most relevant to this paper.

\subsection{Multi-sensor Fusion for SLAM}

\subsubsection{LiDAR-Inertial-Camera Fusion for SLAM}

State estimation in LIC systems has been achieved either by graph-based optimization~\cite{zhang2018laser,wang2019robust,lowe2018complementary,shan2021lvi,zhao2021super,wisth2021unified, wisth2021vilens,jia2021lvio} or filter-based methods~\cite{zuo2019lic,zuo2020lic,lin2021r3live}, while work~\cite{lin2021r} also combines both the graph optimization and filter for localization and mapping. We summarize typical LIC systems in Tab.~\ref{tab:lic_related_works}, which depicts the processing methods of raw sensory measurements as well as the state estimation framework.
Fig.~\ref{fig:four_factor_graph} summarizes four typical pipelines using factor-graph optimization, including the loosely-coupled ones (F1, F2, and F3), as well as the tightly-coupled methods (F4).
V-LOAM~\cite{zhang2014loam} is a loosely-coupled method, comprised of IMU integration, vision and integrated IMU poses fusion with factor-graph optimization, as well as LiDAR and VIO poses fusion with factor-graph optimization (see pipeline F1 in Fig.~\ref{fig:four_factor_graph}). 
All the three submodules can output pose estimation at different frequencies, and the estimated poses are refined step by step within the submodules. 
Some works~\cite{wang2019robust,khattak2020complementary}, utilizing visual-inertial odometry to provide initial pose guess for LiDAR point cloud registration (see F2 of Fig.~\ref{fig:four_factor_graph}), adopt another way to loosely couple sensor measurements. Some other works~\cite{shan2021lvi, zhao2021super} get pose estimation via IMU preintegration, visual-inertial odometry and LiDAR-inertial odometry separately, then fuse the three types of pose estimation in a pose graph, which can be solved by factor-graph optimization. 
In contrast to loosely-coupled methods, which somehow fuse intermediate pose estimation from sensor measurements, tightly coupled approaches~\cite{wisth2021unified,wisth2021vilens,jia2021lvio,lowe2018complementary} directly integrate sensor data, estimating system state with IMU data (preintegration/raw measurements), image features and LiDAR features (illustrated in F4 of Fig.~\ref{fig:four_factor_graph}).

\noindent
\textbf{IMU measurement:} 
Regarding the measurement processing for different sensor modalities, filter-based methods usually use IMU measurements to propagate system states, and optimization-based methods widely adopt preintegration~\cite{forster2016manifold} technique which integrates high-rate IMU data into a low-rate preintegration factor; continuous-time trajectory based methods naturally couple raw IMU measurements in the batch optimization.

\noindent
\textbf{Visual measurement:} 
Visual algorithms could be categorized into the indirect and the direct upon the visual residual models~\cite{engel2017direct}. Indirect methods extract and track features from images and construct geometric constraints in the estimation, while direct methods directly use the actual image values to formulate photometric error, allowing for a more finely grained geometry representation. Both methods rely on accurate depth estimation of landmarks, which can be enhanced by the highly accurate clouds of LiDAR. Specifically, Lowe et al.\cite{lowe2018complementary} set the depth of a feature to the nearest surfel along the feature ray, and set the depth uncertainty to the surfel's covariance in the ray direction. Other methods~\cite{zhang2018laser,shan2021lvi,wisth2021vilens,zhao2021super} first project the LiDAR cloud onto the spherical coordinates of the image, then associate the image feature with the nearest local planar patch that is formed by several nearest LiDAR points, and finally set the feature depth to the depth of the intersection between the ray (from the camera center to the feature) and the plane.

\noindent
\textbf{LiDAR measurement:} 
Line and plane features based on local patch's smoothness firstly defined in LOAM algorithm~\cite{zhang2014loam} are popular in data association of LiDAR clouds, followed by several methods~\cite{zhang2018laser, wang2019robust, shan2021lvi,zuo2019lic}. Zuo et al.~\cite{zuo2020lic} extend to track consecutive plane feature and VILENS~\cite{wisth2021unified,wisth2021vilens} tracks both line and plane feature. 

\subsubsection{Multi-LiDAR Fusion for SLAM}
Multi-LiDAR odometry LOCUS~\cite{LOCUS} and its following~\cite{reinke2022locus2} combine motion-corrected scans from each LiDAR into a single point cloud, which is registrated by Generalized Iterative Closest Point (GICP).
MILIOM~\cite{nguyen2021miliom} chooses to extract features from raw organized scan first and merges feature cloud of each LiDAR instead of merging the full scan, the feature cloud is registrated through feature-to-map matching method.

\subsection{SLAM with Continuous-Time Trajectory}

Continuous-time trajectory representation includes linear interpolation, wavelets, Gaussian process and splines, in this paper we focus on B-spine based representation as explained in Sec.~\ref{sec:bspline_traj}.
Pioneering work on solving B-spline based continuous-time SLAM problem is first systematically derived in ~\cite{furgale2012continuous}, where the authors propose to represent the states as a weighted sum of continuous temporal basis function and illustrate its application case in the extrinsic calibration between IMU and camera.
Thereafter, B-spline based continuous-time trajectory formulation has been widely applied to SLAM-relevant applications, such as event camera odometry~\cite{mueggler2018continuous}, and LiDAR-inertial odometry~\cite{lv2021clins, park2021elasticity}, multi-camera SLAM system~\cite{yang2021asynchronous} and LIC fusion~\cite{lowe2018complementary}. 
Lowe et al.~\cite{lowe2018complementary} tightly couple LIC measurements to estimate state in continuous-time trajectory, however they assume no timeoffset between sensors which does not hold well in real world application, and prior information is discarded without any explanation. 
Our previous work~\cite{lv2021clins} proposes a continuous-time based method for LI system, which manages the measurements within a local window and presents a two-stage loop closure strategy to obtain global-consistent trajectory. Some IMU measurements older than current scan are included in local window rather than applying marginalization technique to retain information about the old measurements, and in addition, it uses automatic derivation to solve the NLS problem. Due to these two aspects, the system is not able to run in real time.

In this paper, we propose a continuous-time based method that supports fusing measurements from LiDAR, IMU and camera, which is very easy to expand to fuse more other sensors to improve the accuracy of localization and mapping, such as fusing GPS in outdoor cases and RFID~\cite{li2021quantized} in indoor cases.

\section{Preliminary on Continuous-time Trajectory}
\label{sec:traj}

This section presents in detail continuous-time trajectory representation based on B-spline and its time derivatives.

\subsection{B-spline Trajectory Representation}
\label{sec:bspline_traj}
We employ B-spline to parameterize continuous-time trajectory as it has the good property of locality and closed-form analytic time derivatives, $C^{k-1}$ smoothness for a spline of order $k$ (degree $k-1$)~\cite{lyche2008spline}.
Specifically, we parameterize the continuous-time 6-DoF trajectory with uniform cumulative B-splines in a split representation format~\cite{haarbach2018survey}. The translation $\bp(t)$ of $k$ order over time $t\in [t_i, t_{i+1})$ is controlled by the temporally uniformly distributed translational control points $\bp_i$, $\bp_{i+1}$, $\dots$, $\bp_{i+k-1}$, and the matrix format~\cite{qin1998general} could be written as:
\begin{align}
    \label{Eq:p_curve}
    \underbrace{\bp(t)}_{3\times 1} = 
    \underbrace{\begin{bmatrix} \bp_i & \bd_1^i & \cdots & \bd_{k-1}^i \end{bmatrix}}_{3\times k} \underbrace{\widetilde{\bM}^{(k)}}_{k\times k}\underbrace{\bu}_{k\times 1} \\
    \bu = \begin{bmatrix} 1 & u & \cdots & u^{k-1} \end{bmatrix}, u=(t-t_i)/(t_{i+1}-t_i) \notag
\end{align}
with difference vectors $\bd_j^i = \bp_{i+j}-\bp_{i+j-1}\in \R^3$.
The cumulative spline matrix $\widetilde{\bM}^{(k)}$ of uniform B-spline only depends on the B-spline order. We further define $\blambda(t) = \widetilde{\bM}^{(k)} \bu$ thus Eq.~\eqref{Eq:p_curve} can be written as 
\begin{align}
    \bp(t) = \bp_i + \sum_{j=1}^{k-1}{\lambda_j(t)\cdot\bd^i_j}\,.
\end{align}
To parameterize the 3D rotation in $\SO(3)$, we adopt the cumulative B-splines in Lie groups with the following expression over time $t\in [t_i, t_{i+1})$: 
\begin{align}
    \label{Eq:R_curve}
    \bR(t) = \bR_i \cdot \prod_{j=1}^{k-1}{\Exp\left(\lambda_j(t)\cdot\Log\left(\bR_{i+j-1}^{-1}\bR_{i+j}\right)\right)}
\end{align}
where $\bR_i\in\SO(3)$ are the control points for rotation. The difference vector between two rotations is defined as $\bd^i_j = \Log\left(\bR_{i+j-1}^{-1}\bR_{i+j}\right) \in \R^3$ 
and $\bA_j(t) = \Exp\left(\lambda_j(t)\cdot\bd_j\right)$ where omitting the $i$ to simplify notation, Eq.~\eqref{Eq:R_curve} can be written in the following concise equation:
\begin{align}
    \bR(t) = \bR_i \cdot \prod_{j=1}^{k-1}{\bA_j(t)}\,.
\end{align}
In this paper, we select cubic (degree = 3) B-spline and the corresponding cumulative spline matrix $\widetilde{\bM}^{(k)}$ is
\begin{align}
    \widetilde{\bM}^{(4)} =  \frac{1}{6}
    \begin{bmatrix} 6 & 5 & 1 & 0 \\
    0 & 3 & 3 & 0 \\
    0 & -3 & 3 & 0 \\
    0 & 1 & -2 & 1 \\
    \end{bmatrix}
    \text{.}
\end{align}

\subsection{Time Derivatives of B-spline}
\label{sec:traj_time_derivative}
As mentioned before, B-spline provides closed-form analytic derivatives, enabling the proposed system to fuse high-frequency IMU measurements seamlessly. Continuous-time trajectory of IMU in global frame $\{G\}$ is denoted as ${}^G_I\bT(t) = \left[{}^G_I\bR(t), {}^G\bp_I(t)\right]$, and its time derivatives can be derived:
\begin{align}
    \label{eq:spline_dp}
    {}^G\bv(t) &= {}^G\dotbp_I(t) = \sum_{j=1}^{3}{\dot{\lambda}_j(t)\cdot\bd^i_j} \text{,}  \\
    \label{eq:spline_ddp}
    {}^G\ba(t) &= {}^G\ddotbp_I(t) = \sum_{j=1}^{3}{\ddot{\lambda}_j(t)\cdot\bd^i_j} \text{,} \\
    \label{eq:spline_dR}
    {}_I^G\dotbR(t) &= \bR_i\left(\dot{\bA}_1 \bA_2 \bA_3 + \bA_1 \dot{\bA}_2 \bA_3 + \bA_1 \bA_2 \dot{\bA}_3\right)
\end{align}
where $\dot{\bA}_j = \Exp\left(\dot{\lambda}_j(t)\cdot\bd_j\right)$. Naturally, it's straightforward to compute the linear accelerations and angular velocities in local IMU frame:
\begin{align}
\label{eq:derivative}
    {}^I\ba(t) &=
    {}_I^G\bR^\top(t) \left( {}^G\ba(t) - {}^G\bg \right) \\
    {}^I\bomega(t) &= {}_I^G\bR^\top(t) \cdot {}_I^G\dotbR(t)
\end{align}
where $^G\bg \in \R^3$ denotes the gravity vector in global frame.

In the following section, we model the continuous-time trajectory of IMU sensor in $\{G\}$ frame, termed as ${}^G_I\bT(t)$. The global frame $\{G\}$ is determined by the first IMU measurement after system initialization by aligning its z-axis with the gravity direction,  and the gravity can be denoted as $[0,0,9.8]$ in $\{G\}$. 
We assume the extrinsic rigid transformations between sensors are precalibrated~\cite{lv2022observability}, and we can get the camera trajectory ${}^G_C\bT(t)$, and LiDAR trajectory ${}^G_L\bT(t)$ handily by transferring IMU trajectory ${}^G_I\bT(t)$ with the known extrinsic transformations.
\section{Continuous-Time Fixed-Lag Smoothing}
\label{sec:fixed-lag-smoothing}

\begin{table}[t]
    \caption{Notations Glossary}
    \centering
    \begin{tabular}{@{}ll@{}}
        \toprule
        Symbols & Meaning \\ \midrule
         $\bPhi(t_{\kappa-1}, t_{\kappa})$  & control points of B-splines in $\left[ t_{\kappa-1}, t_{\kappa}\right)$  \\
         $\bPhi_R(t_{\kappa}) / \bPhi_p(t_{\kappa})$  & involved orientation/position control points at $ t_{\kappa}$  \\
         $\bb_{\omega}^{\kappa}, \bb_{a}^{\kappa}$ & biases of temporal sliding window in $\left[ t_{\kappa-1}, t_{\kappa}\right)$ \\ 
         $ t_L, t_I, t_C$ & timeoffsets of LiDAR, IMU and camera, respectively \\
         $t$ & timestamp of trajectory or sensor measurements \\
         $\tau = t + t_{\text{offset}}$ & corrected timestamp of sensor measurements \\
         \bottomrule
    \end{tabular}
    \label{tab:notation}
\end{table}

\begin{figure}[t]
    \centering
    \includegraphics[width=1\linewidth]{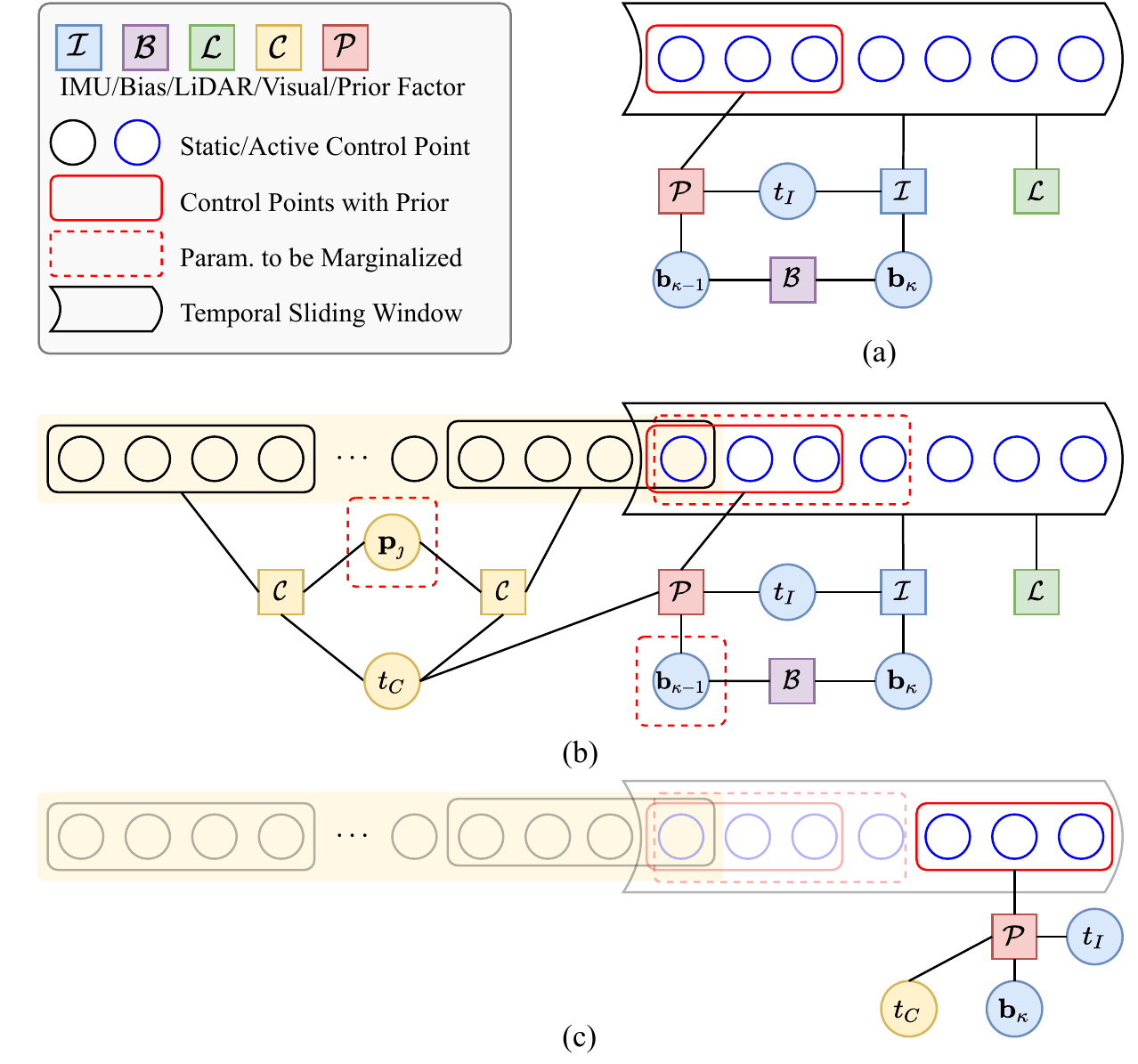}
    \captionsetup{font={small}}
    \caption{Factor graphs of multi-sensor fusion. (a) A typical factor graph of LI system fusion. (b) A typical factor graph of LIC system fusion. Active control points are to be optimized, while static control points remain constant. The control points with yellow background are involved in visual keyframe sliding window.
    (c) After marginalization of \textit{(b)}, the induced prior factor is involved with the latest control points, latest bias and timeoffsets.}
    \label{fig:clic_factor_graph}
\end{figure}

This section presents the continuous-time fixed-lag smoothing applied in factor-graph optimization, which is the core of estimator design. Before diving into details, we introduce some notations used in this paper which is summarized in Tab.~\ref{tab:notation}.  

\subsection{Factor-Graph Optimization}
\label{sec:factor_graph}
We fuse heterogeneous IMU, LiDAR, and camera measurements in a factor graph optimization framework with the continuous-time trajectory formulation. 
Figure.~\ref{fig:clic_factor_graph} (a) and (b) show the factor graphs of the LI system and LIC system, respectively. Here we only illustrate the factor graph for LIC system, and it is straightforward to apply to LI system with minor adaptions. Specifically, our estimator in the temporal sliding window (detailed in Sec.~\ref{sec:temporal_win}) over $\left[ t_{\kappa-1}, t_{\kappa}\right)$ aims to estimate the following states:
\begin{align}
    \label{eq:state}
	\mathcal{X}^{\kappa} &= \{\bx_R^{\kappa},\, \bx_p^{\kappa},\, \bx_{I_b}^{\kappa},\, \bx^{\kappa}_{\lambda},\, t_{I},\, t_{C}\}\,, \\
	\bx_{I_b}^{\kappa} &= \{ \bb_{\omega}^{\kappa-1},\, \bb_{a}^{\kappa-1},\, \bb_{\omega}^{\kappa},\, \bb_{a}^{\kappa} \} \,, \notag
\end{align}
which include active control points of B-splines $\bPhi(t_{\kappa-1}, t_{\kappa})=\{\bx_R^{\kappa}, \bx_p^{\kappa}\}$, the IMU biases $\bx_{I_b}^{\kappa}$, the parameters of visual landmarks $\bx_{\lambda}^{\kappa}$, and the temporal offset between LiDAR and IMU $t_{I}$ or camera $t_{C}$. Notably, LiDAR is taken as the base sensor in our multi-sensor fusion system, thus we need to align IMU and camera timestamps to LiDAR. Sec.~\ref{sec:timeoffset_estimate} provides more details about the timeoffset estimation.

The factor graph needed to be solved consists of LiDAR factors $\br_L$, IMU factors $\br_I$, one bias factor $\br_{I_b}$, visual factors $\br_C$, and one prior factor $\br_{\text{prior}}$ induced from marginalization, the details of factors are provided in the following sections.
With the LiDAR-inertial measurements during $\left[ t_{\kappa-1}, t_{\kappa}\right)$ and  tracked visual features in the keyframe sliding window (detailed in Sec.~\ref{sec:keyframe_sliding_win}), we formulate the following \textbf{N}onlinear \textbf{L}east-\textbf{S}quares (\textbf{NLS}) problem:
\begin{align}
  \label{eq:least-squares}
  \hat{\mathcal{X}}^{\kappa} = \underset{\mathcal{X}^{\kappa}}{\operatorname{argmin}} \;\br, \;\;  \br= \br_I + \br_{I_b} + \br_L + \br_c + \br_{\text{prior}} \text{,}
\end{align}
and solve the NLS problem by iterative optimization methods. 
At each iteration, the system is linearized at current estimate $\hat{\bx}$, and we define its  error state as $\tilde{\bx} = \bx - \hat{\bx}$, where $\bx$ is the true state.
In practice we adopt the Levenberg-Marquardt algorithm from the Ceres Solver~\cite{Agarwal_Ceres_Solver_2022} library and employ analytical derivatives to speed up the NLS problem-solving. 

\subsection{LiDAR Factor}
A LiDAR point measurement ${}^L\bp_t$ with noise $\bn_L$, measured at time $t_{\ell}$, is associated with a 3D plane in closest point parameterization~\cite{geneva2018lips, yang2019tightly}, ${}^{G}\bpi = {}^{G}d_{\pi}{}^{G}\bn_{\pi}$, where ${}^{G}d_{\pi}$ and ${}^{G}\bn_{\pi}$ denote the distance of the plane to origin and unit normal vector, respectively. We can transform LiDAR point to global frame by  
\begin{align}
	{}^{G}\hat{\bp}_{\ell} = {}^G_L\bR(\tau_{\ell}) \left( {}^L\bp_{\ell} + \bn_L \right) + {}^G\bp_L(\tau_{\ell}) 	\text{,}
\end{align}
and the point-to-plane distance is given by:
\begin{equation}
\begin{aligned}
    \label{eq:lidar_factor}
	\br_{L}(\tau_{\ell}, \hat{\mathcal{X}^{\kappa}}, {}^L\bp_{\ell}, {}^{G}\bpi) &= {}^{G}\bn^\top_{\pi} {}^{G}\hat{\bp}_{\ell} + {}^{G}d_{\pi} \\
    &\approx \bH_{\ell} \cdot \tilde{\bx} + \bG_{\ell}\cdot \bn_{L} \text{.}
\end{aligned}
\end{equation}
where $\tau_{\ell} = t_{\ell}$ is the measure time of LiDAR point, $\bH_{\ell}$ is Jacobian matrix w.r.t error state $\tilde{\bx}$ 
(see our supplementary file for detail). 
$\bn_L$ is assumed to be under independent and identically distributed (i.i.d.) white Gaussian noise in our experiments. $\bG_{\ell}$ is the Jacobian with respect to $\bn_L$ which can be easily computed.

\subsection{IMU Factor and Bias Factor}
Considering raw IMU measurements at $t_m$ with angular velocity ${}^I\bomega_m$ and linear acceleration ${}^I\ba_m$, and the true angular velocity and linear acceleration are denoted by ${}^I\bomega$ and ${}^I\ba$, respectively. The following equations hold:
\begin{align}
    &{}^I\bomega_m = {}^I\bomega(t) + \bb_{\omega}(t) + \bn_{\omega} \\
    &{}^I\ba_m(t) = {}_I^G\bR^\top(t) \left( {}^G\ba(t) - {}^G\bg \right) + \bb_{a}(t) + \bn_{a} \\
    &\dot{\bb}_{\omega}(t) = \bn_{b_\omega},\quad \dot{\bb}_{a}(t) = \bn_{b_a}
\end{align}
where $\bn_{\omega},\,\, \bn_{a}$ are zero-mean Gaussian white noise. The gyroscope bias $\bb_{\omega}$ and accelerometer bias $\bb_{a}$ are modeled as random walks, driving by the white Gaussian noises $\bn_{b_\omega}$ and $\bn_{b_a}$, respectively. 
With the timeoffset $t_I$ of IMU between LiDAR, we have the following IMU factor
\begin{equation}
\begin{aligned}
    \label{eq:imu_factor}
    &\br_{I}(\tau_m, \hat{\mathcal{X}^{\kappa}}, {}^I\bomega_m, {}^I\ba_m) \\
    &= \begin{bmatrix} 
    {}^I\bomega(\tau_m) - {}^I\bomega_m + \bb_{\omega}^{\kappa} \\
    {}^I\ba(\tau_m) - {}^I\ba_m + \bb_a^{\kappa}
    \end{bmatrix} 
    + \begin{bmatrix} 
    \bn_{\omega} \\
    \bn_{a}
    \end{bmatrix} \\
    &\approx \bH_{I_m}^{\kappa}\cdot \tilde{\bx} + \bG_{I_m} \cdot\bn_I \text{,}
\end{aligned}
\end{equation}
and bias factor
\begin{equation}
\begin{aligned}
    \label{eq:bias_factor}
    \br_{I_b}(\hat{\mathcal{X}}^{\kappa}) &= \begin{bmatrix} 
    \bb_{\omega}^{\kappa} - \bb_{\omega}^{\kappa-1} \\
    \bb_{a}^{\kappa} - \bb_{a}^{\kappa-1}
    \end{bmatrix} 
    + \begin{bmatrix} 
    \bn_{b_\omega} \\
    \bn_{b_a}
    \end{bmatrix} \\
    &\approx \bH_{I_b}^{\kappa}\cdot \tilde{\bx} + \bG_{I_b} \cdot \bn_{I_b} 
\end{aligned}
\end{equation}
where $\tau_m = t_m + t_I$ is the corrected IMU timestamp, 
and $\bH_{I_m}^{\kappa}, \bH_{I_b}^{\kappa}$ are Jacobian matrices with respect to states 
(see supplementary file), 
and $\bG_{I_m}, \bG_{I_b}$ are Jacobian matrices with respect to noise. By substituting the derivative of continuous-time trajectory (Eq.~\eqref{eq:derivative}) at time instant $\tau_m$ into Eq.~\eqref{eq:imu_factor}, we can optimize the continuous-time trajectory by raw IMU measurements directly, avoiding the efforts of IMU propagation or pre-integration.

\subsection{Visual Factor}
A landmark $\bp_\jmath$ observed in its anchor keyframe $\mathcal{F}_a$ at timestamp $t_a$ and observed again in frame $\mathcal{F}_b$ at timestamp $t_b$, can be given the initialized inverse depth as $\lambda_\jmath$ through triangulation (as Sec.~\ref{sec:visual_frontend}).  Let $\brho_{\jmath}^{a}$ denotes 2D raw observation in $\mathcal{F}_a$ with noise $\bn_c$, the estimated position of landmark in frame $\mathcal{F}_b$ is
\begin{align} 
    \hat{\bp}_{{\jmath}}^{b} = {}^G_C\bT(\tau_{b})^{\top} \cdot {}^G_C\bT(\tau_{a})\cdot \frac{1}{\lambda_{\jmath}} \pi_c( \brho_{\jmath}^{a}  + \bn_c)
\end{align}
where $\pi_c(\cdot)$ denotes the back projection which transforms a pixel to the normalized image plane. $\tau_a, \tau_b$ are corrected timestamps to remove timeoffsets. 
The corresponding visual factor based on reprojection error is defined as:
\begin{equation}
\begin{aligned}
    \label{eq:visual_factor}
    \br_{c}(\tau_{b}, \hat{\mathcal{X}^{\kappa}}, \brho_{\jmath}^{b}) &= 
    \begin{bmatrix} \be_1^{\top} \\ \be_2^{\top} \end{bmatrix} \left(
    \frac{\hat{\bp}_{{\jmath}}^{b}}{\be_{3}^{\top} \hat{\bp}_{{\jmath}}^{b}} - \pi_c\left( \brho_{\jmath}^{b} + \bn_c \right) \right) \\
    &\approx \bH_{\jmath}^{b} \cdot \tilde{\bx} + \bG_{\jmath}^{b} \cdot \bn_c
\end{aligned}
\end{equation}
where $\be_{i}$ denotes a $3\times1$ vector with its $i$-th element to be 1 and the others to be 0, and $\brho_{\jmath}^{b}$ describes 2D raw observation in $\mathcal{F}_b$. $\bH_{\jmath}^{b}$ denotes Jacobian matrix 
(see supplementary file).

\subsection{Marginalization}
\label{sec:marg}
To bound the size of sliding window in our continuous-time fixed-lag smoother, marginalization has to be resorted to. Although marginalization is universally leveraged in discrete-time sliding-window estimator~\cite{leutenegger2015keyframe,qin2018vins}, it is rarely investigated in continuous-time sliding-window estimator. By utilizing probabilistic marginalization, information on the active states can be well reserved in the estimator, when measurements and old states are removed from the sliding window.
Linearizing the factors at the best estimate of the state gives:
\begin{align}
    \begin{bmatrix}
    \bH_{\alpha\alpha} & \bH_{\alpha\beta} \\
    \bH_{\beta\alpha} & \bH_{\beta\beta}
    \end{bmatrix}  
    \begin{bmatrix}
    \bx_\alpha \\ \bx_\beta
    \end{bmatrix} = 
    \begin{bmatrix}
    \bb_\alpha \\ \bb_\beta
    \end{bmatrix}
\end{align}
where we organize the reserved parameters in $\bx_\alpha$, and $\bx_\beta$ will be marginalized. Using the Schur-Complement~\cite{sibley2010sliding}, the following equation holds:
\begin{align}
    \left(\bH_{\alpha\alpha} - \bH_{\alpha\beta}\bH_{\beta\beta}^{-1}\bH_{\beta\alpha}\right)\bx_\alpha &= \left(\bx_\alpha - \bH_{\alpha\beta}\bH_{\beta\beta}^{-1}\bx_\beta\right)
\end{align}
 and by introducing new notations, we can denote the above equation by:
\begin{align}
    \hat{\bH}_{\alpha\alpha} \bx_\alpha = \hat{\bb}_\alpha
    \label{eq:marg_factor}
\end{align}
which is exactly the prior factor. 
Figure.~\ref{fig:clic_factor_graph} (b) displays the entries needed to be marginalized out of the LI temporal- and visual keyframe-sliding windows. The entries needed to be marginalized vary at different statuses. We marginalize out the control points and IMU biases when sliding the LI temporal window, and marginalize out the inverse depths of landmarks when sliding the oldest keyframe out of visual keyframe sliding window, producing a prior on 
\begin{align}
    \mathcal{X}_{prior}^{\kappa} = \left\{ \bPhi(t_{\kappa-1}, t_{\kappa}) \cap \bPhi(t_{\kappa}, t_{\kappa+1}), \, \bb_{\omega}^{\kappa}, \bb_{a}^{\kappa}, t_I,t_C \right\} \text{.} \notag
\end{align}
where $\bPhi(t_{\kappa-1}, t_{\kappa}) \cap \bPhi(t_{\kappa}, t_{\kappa+1})$ denotes the affected control points in next sliding window. 
The prior factor induced from marginalization is shown in Fig.~\ref{fig:clic_factor_graph} (c) and will be involved in future factor-graph optimization.

\section{Multi-sensor Fusion}
\label{sec:method}

\begin{figure}[t]
    \centering
    \includegraphics[width=1\linewidth]{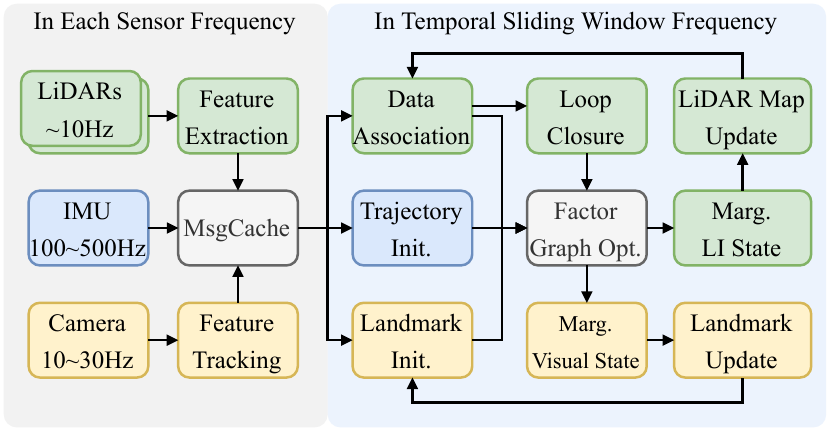}
    \captionsetup{font={small}}
    \caption{The pipeline of the proposed LiDAR-Inertial-Camera fusion system. Raw IMU measurements, features of each LiDAR and tracked features of camera are cached in the MsgCache module, and measurements are fed to the sliding window at  $\frac{1}{\eta \Delta t}$ Hz. After factor graph optimization, we separately marginalize LI state and visual state, and update local LiDAR map and visual landmarks.
    More details are provided in Sec.~\ref{sec:method}. }
    \label{fig:pipline}
\end{figure}

This section presents in detail the multi-sensor fusion method powered by the versatile continuous-time fixed-lag smoothing presented above. Fig.~\ref{fig:pipline} shows the overall architecture of a LIC fusion system.
In this section, we first introduce the LiDAR-IMU system (Sec.~\ref{sec:lio_sys}) and visual system (Sec.~\ref{sec:visual_sys}). Then we reveal some implementation details (Sec.~\ref{sec:implementation_detail}) to accommodate the particularity of  continuous-time trajectory estimation, which is with significant distinction from discrete-time methods.

\subsection{LiDAR-Inertial System}
\label{sec:lio_sys}

\subsubsection{LiDAR Measurement Processing}
There are three main stages for LiDAR point cloud processing: feature extraction, data association, and local map management.
Inspired by ~\cite{zhang2014loam}, for each new incoming LiDAR scan,  we first compute the curvature of each point according to its neighboring points, and select points with small curvature as planar points.
The motion distortion in raw LiDAR scan is inevitable due to the motion when the LiDAR is scanning. Since we have formulated the continuous-time trajectory, it is handy to query poses at every time instant, which allows us to compensate for the motion distortion by aligning all the LiDAR points in one scan to the sweeping start time of that scan.
After removing the incidental motion distortion in raw LiDAR scan and projecting undistorted LiDAR scan to the map frame using the initialized (or optimized) continuous-time trajectory, we can perform the point-to-plane data association by associating planar LiDAR points to tiny planes in the map. The tiny planes are found by fitting neighboring planar points on the map. The point-to-plane distance (see Eq.~\eqref{eq:lidar_factor}) will be minimized in the continuous-time fixed-lag smoother.  After finishing the first optimization, we update the data association using the optimized trajectory and optimize again to refine the estimation.
Upon completion of the optimization, we individually transform each LiDAR point into the scan's start time according to queried pose from the optimized trajectory, and select keyscans based on the displacement of poses or time span. Keyscans are leveraged to build the local LiDAR map in $\{G\}$ frame, and the local LiDAR map comprised of certain keyscans keeps updated upon newly-added scan.

\subsubsection{LI Temporal Sliding Window}
\label{sec:temporal_win}
For the LI system, we maintain a temporal sliding window within a constant time duration, $\eta \Delta t$, where $\Delta t$ denotes the B-spline temporal knot distance and $\eta$ is an integer. The continuous-time trajectory of LI system is optimized and updated every $\eta \Delta t$ seconds. 
Compared to optimizing the trajectory within every LiDAR scan individually~\cite{lv2021clins}, the temporal sliding-window optimization (fixed-lag smoothing) makes full use of all measurements within the sliding window (see Fig.~\ref{fig:timeline}) and prior information from marginalization, which could achieve higher accuracy.
Note that the IMU bias sampling frequency is set as $\frac{1}{\eta \Delta t}$Hz, that is to say, we add a new IMU bias state when sliding the temporal window forward, and the discrete noise of bias factor (see Eq.~\eqref{eq:bias_factor}) is determined accordingly.

\subsection{Visual System}
\label{sec:visual_sys}
The main purpose of incorporating camera to the multi-sensor fusion estimator is to improve the robustness of LI system. We expect the LiDAR-Inertial-Camera system not to have degraded performance in structureless scenarios, which is a significant challenge for LI systems.

\subsubsection{Visual Front End}
\label{sec:visual_frontend}
We extract corner features~\cite{shi1994good} from the images with KLT tracking~\cite{lucas1981iterative} and determine image keyframe based on parallax variation and the number of features tracked, similar to~\cite{qin2018vins}.
Furthermore, we triangulate the landmarks tracked throughout the whole keyframe sliding window (see Sec.~\ref{sec:keyframe_sliding_win}) using keyframe poses queried from current best-estimated continuous-time trajectory, and only keep the landmarks with low reprojection errors. Landmarks are parameterized by inverse depth represented in anchor frame, which in our case is always the oldest keyframe in the keyframe sliding window.

\subsubsection{Visual Keyframe Sliding Window}
\label{sec:keyframe_sliding_win}
\begin{figure}[t]
    \centering
    \includegraphics[width=1\linewidth]{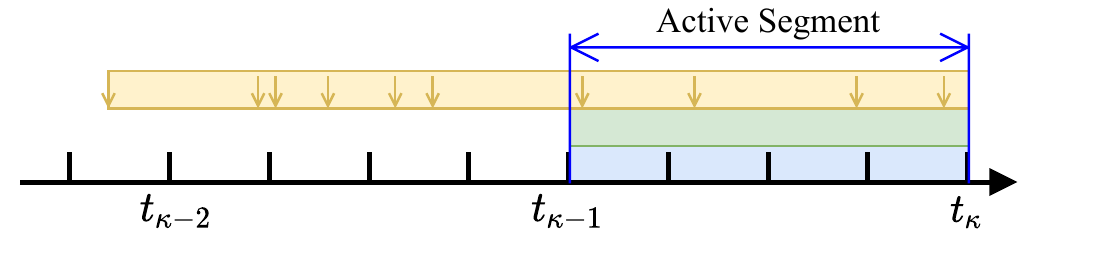}
    \vspace{-1em}
    \captionsetup{font={small}}
    \caption{
    The involved measurements from IMU (blue block) and LiADR (green block) sensors in a temporal sliding window in $\left[t_{\kappa-1}, t_{\kappa}\right)$, and measurements of camera (yellow block, keyframes denoted by arrows) in a keyframe sliding window. The active trajectory segment of which control points will be optimized is within the LI temporal sliding window.}
    \label{fig:timeline}
\end{figure}

When processing images, we maintain a visual keyframe sliding window with a constant number of keyframes, in contrast to the constant time duration for LI temporal sliding window. This is due to the consideration that visual landmark triangulation needs observations across multiple keyframes with sufficient parallaxes. 
The duration of the keyframe sliding window is normally longer than the LI temporal sliding window, and estimating the entire trajectory covered by all the keyframes is time-consuming. 
In practice, we determine the trajectory optimization range based on the temporal sliding window of the LI system and define the control points to be optimized as active control points (shown in Fig.~\ref{fig:timeline}). We only optimize the control points of active trajectory segment within the LI temporal sliding window, while the other control points are involved in the optimization but kept static in the optimization. That formulation of visual system may not the best choice of high-accuracy estimation, but rather a trade-off to achieve real-time.
While for the marginalization of visual keyframe sliding window, the oldest keyframe and the landmarks in it are marginalized out.

The strategy of sliding the keyframe window is similar to~\cite{qin2018vins}, where the newest frame in the sliding window is always the latest image of the system. If the second newest frame is determined as keyframe, we will marginalize out the oldest keyframe and  landmarks in it from the sliding window, in order to keep a constant number of keyframes.
Otherwise, we discard the second newest frame directly.
Note that, we need to transfer the anchor frame of visual landmark if its anchor frame is marginalized.

\begin{figure}[t]
    \centering
    \includegraphics[width=1\linewidth]{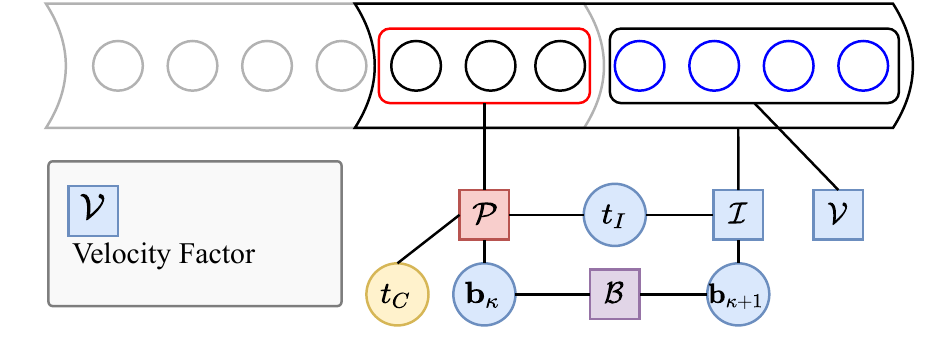}
    \captionsetup{font={small}}
    \caption{A factor graph of trajectory initialization with details provided in Sec.~\ref{sec:initialization}.}
    \label{fig:clic_init_graph}
\end{figure}

\subsection{Extra Implementation Details}
\label{sec:implementation_detail}

\subsubsection{Initialization} 
\label{sec:initialization}

We initialize the IMU bias and gravity direction using the raw IMU measurements, assuming that the system is stationary at the start, which shares the same spirit as~\cite{Geneva2020ICRA}.
The system appends new IMU biases and control points when sliding the LI temporal window. The new IMU biases are initialized to the value of the previous temporal sliding window bias, and the new control points are first assigned values of the neighboring control point and further initialized via factor graph optimization as shown in Fig.~\ref{fig:clic_init_graph}. The velocity factor is defined as 
\begin{align}
    \br_{v} = {}^G\hat{\bv}_t - {}^G\bv(t)\,, 
\end{align}
where ${}^G\bv(t)$ is defined in Eq.~\eqref{eq:spline_ddp}, 
which is derivative of continuous-time trajectory. ${}^G\hat{\bv}_t$ is an estimate of global linear velocity at the end of current LI temporal window, and is derived by forward integrating IMU measurements from the pose at the start of temporal window. 
When solving the initialization problem, only the newly added control points are optimized while all the other states remain constant during optimization. 
After initialization, we remove the distortion of LiDAR scans with that initialized trajectory for better association results.

\subsubsection{Online Calibration of Timeoffset}
\label{sec:timeoffset_estimate}
Estimating timeoffsets between sensors is natural and convenient when modeling the trajectory in continuous time. In this paper, we choose the LiDAR sensor as the time baseline. Thus the timeoffsets of the camera and IMU with respect to the LiDAR need to be known or calibrated online. 
The main reason of choosing LiDAR as the base sensor is that local LiDAR map needs to be maintained, and once the timeoffset of LiDAR trajectory is changed, LiDAR map needs to be transformed accordingly, which is time-consuming.
In contrast, the extra computation arising from the change of camera or IMU timestamps is insignificant.

\subsubsection{Loop Closure}

We utilize Euclidean distance-based loop closure detection method~\cite{shan2020lio} and adopt the two-stage continuous-time trajectory correction method~\cite{lv2021clins} to tackle loop closures. After a loop closure optimization, we will remove the prior information of states since the current best-estimated states may be away from the linearized points, resulting in inappropriate prior constraints.

\section{Experiment}
\label{sec:experiment}

\begin{table*}[t]
	\centering
	\captionsetup{font={small}}
	\caption{The APE (RMSE, meter) results on VIRAL dataset. The best result is in bold, and the second best is \underline{underlined}.
	}
	\begin{tabular}{@{}lccccccccccr@{}}
	\toprule
    Method & Sensor${}^{(1)}$ & \begin{tabular}[c]{@{}c@{}}eee\_01\\ (237m)\end{tabular} & \begin{tabular}[c]{@{}c@{}}eee\_02\\ (171m)\end{tabular} & \begin{tabular}[c]{@{}c@{}}eee\_03\\ (128m)\end{tabular} & \begin{tabular}[c]{@{}c@{}}nya\_01\\ (160m)\end{tabular} & \begin{tabular}[c]{@{}c@{}}nya\_02\\ (249m)\end{tabular} & \begin{tabular}[c]{@{}c@{}}nya\_03\\ (315m)\end{tabular} & \begin{tabular}[c]{@{}c@{}}sbs\_01\\ (202m)\end{tabular} & \begin{tabular}[c]{@{}c@{}}sbs\_02\\ (184m)\end{tabular} & \begin{tabular}[c]{@{}c@{}}sbs\_03\\ (199m)\end{tabular} & average \\ \midrule
    LIO-SAM${}^{\footnotesize (2)}$~\cite{shan2020lio} & L, I & 0.075 & 0.069 & 0.101 & 0.076 & 0.090 & 0.137 & 0.089 & 0.083 & 0.140 & 0.096 \\
    MILIOM (horz. LiDAR)${}^{\footnotesize (2)}$~\cite{nguyen2021miliom} & L, I & 0.104 & 0.065 & 0.063 & 0.083 & 0.072 & 0.058 & 0.076 & 0.081 & 0.088 & 0.077 \\
    VIRAL (horz. LiDAR)${}^{\footnotesize(2)}$~\cite{nguyen2021viral} & L, I & 0.064 & 0.051 & 0.060 & 0.063 & \underline{0.042} & \underline{ 0.039} & 0.051 & 0.056 & 0.060 & 0.054 \\
    CLINS (w/o loop)~\cite{lv2021clins} & L, I & 0.059 & 0.030 & 0.029 & \textbf{0.034} & \textbf{0.040} & \textbf{0.039} & 0.029 & \underline{0.031} & 0.033 & 0.036 \\
    CLIO (w/o loop) & L, I & \textbf{0.030} & \textbf{0.023} & \underline{ 0.028} & 0.042 & {\color[HTML]{333333} 0.053} & 0.042 & \textbf{0.028} & 0.032 & \textbf{0.030} & \textbf{0.034} \\
    CLIC (w/o loop) & L, I, C & \underline{0.030} & \underline{0.029} & \textbf{0.028} & \underline{0.040} & 0.054 & 0.041 & \underline{0.029} & \textbf{0.031} & \underline{0.033} & \underline{0.035} \\ \midrule
    MILIOM (2 LiDARs)${}^{\footnotesize(2)}$~\cite{nguyen2021miliom} & L2, I & 0.067 & 0.066 & 0.052 & 0.057 & 0.067 & 0.042 & 0.066 & 0.082 & 0.093 & 0.066 \\
    VIRAL (2 LiDARs)${}^{\footnotesize(2)}$~\cite{nguyen2021viral} & L2, I, C & 0.060 & 0.058 & 0.037 & 0.051 & 0.043 & \textbf{0.032} & 0.048 & 0.062 & 0.054 & 0.049 \\
    CLIO2 (w/o loop) & L2, I & \underline{0.040} & \textbf{0.021} & \underline{0.031} & \underline{0.030} & \underline{0.037} & \underline{0.034} & \textbf{0.033} & \underline{0.037} & \underline{0.044} & \underline{0.034} \\
    CLIC2 (w/o loop) & L2, I, C & \textbf{0.038} & \underline{0.025} & \textbf{0.030} & \textbf{0.029} & \textbf{0.036} & 0.035 & \underline{0.034} & \textbf{0.035} & \textbf{0.043} & \textbf{0.034} \\ \bottomrule
	\multicolumn{12}{l}{${}^{(1)}$ Sensors L,I,C are abbreviations of LiDAR, IMU, camera, respectively, L2 represents using two LiDAR sensors.} \\
	\multicolumn{12}{l}{${}^{(2)}$ Results are from~\cite{nguyen2021viral}. The horz. LiDAR in table means only the horizontal LiDAR is used for odometry.}
	\\
	\end{tabular}
	\label{tab:viral_ape}
	\vspace{-1em}
\end{table*}
	
\begin{table}[t]
	\centering
	\captionsetup{font={small}}
	\caption{The APE (RMSE, meter) results on NCD dataset.}
	\resizebox{\linewidth}{!}{
	\begin{tabular}{@{}rccc@{}}
	\toprule
	\multicolumn{1}{c}{Method} & \begin{tabular}[c]{@{}c@{}}NCD\_01\\ (1530s / 1609m)\end{tabular} & \begin{tabular}[c]{@{}c@{}}NCD\_02\\ (2656s / 3063m)\end{tabular} & \begin{tabular}[c]{@{}c@{}}NCD\_06\\ (120s / 97m)\end{tabular} \\ \midrule
	LIO-SAM (w/o loop) & 1.660 & \textbf{2.305} & 0.272 \\
	CLIO (w/o loop) & \textbf{0.792} & 2.686 & \textbf{0.091} \\ \midrule
	LIO-SAM (w/ loop) & 0.544 & 0.592 & 0.272 \\
	CLIO (w/ loop) & \textbf{0.408} & \textbf{0.381} & \textbf{0.091} \\ \bottomrule
	\end{tabular}
	}
	\label{tab:ncd_ape}
	\vspace{-1em}
\end{table}
	
\begin{table}[t]
	\centering
	\captionsetup{font={small}}
	\caption{The APE (RMSE, meter) results on LVI-SAM dataset. }
	\begin{tabular}{@{}rccc@{}} \toprule
	\multicolumn{1}{c}{Method} & Sensor & Handheld (1642s) & Jackal (2182s) \\ \midrule
	LIO-SAM (w/o loop) & L, I & 53.62 & 3.54 \\
	CLIO (w/o loop) & L, I & fail & 3.43 \\
	LVI-SAM (w/o loop) & L, I, C & 7.87 & 4.05 \\
	CLIC (w/o loop) & L, I, C & \textbf{2.56} & \textbf{2.55} \\ \midrule
	LIO-SAM (w/ loop) & L, I & fail & 1.52 \\
	CLIO (w/ loop) & L, I & fail & 1.01 \\
	LVI-SAM (w/ loop) & L, I, C & 0.83 & \textbf{0.67} \\
	CLIC (w/ loop) & L, I, C & 0.65 & 0.88 \\
	CLIC (w/ loop, w/ calib) & L, I, C & \textbf{0.56} & 0.84 \\ \bottomrule
	\end{tabular}
	\label{tab:lvi_ape}
\end{table}

\begin{figure}[t]
	\centering
    \includegraphics[width=1\linewidth]{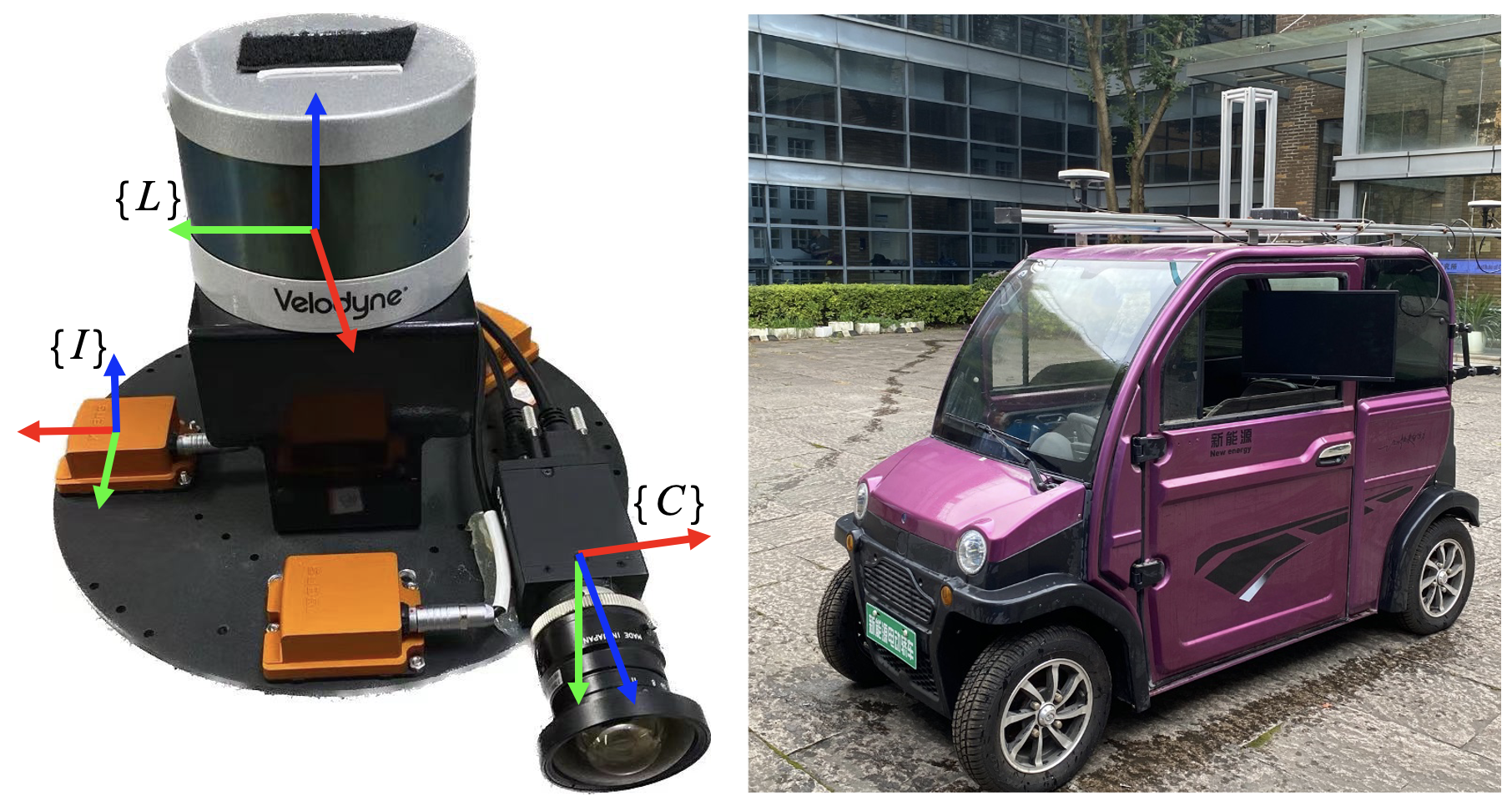}
    \vspace{-1em}
    \captionsetup{font={small}}
	\caption{The sensor rig and ground vehicle to collect YQ dataset.}
	\label{fig:lic_sensors}
\end{figure}

\begin{figure}[t]
	\centering
    \includegraphics[width=1\linewidth]{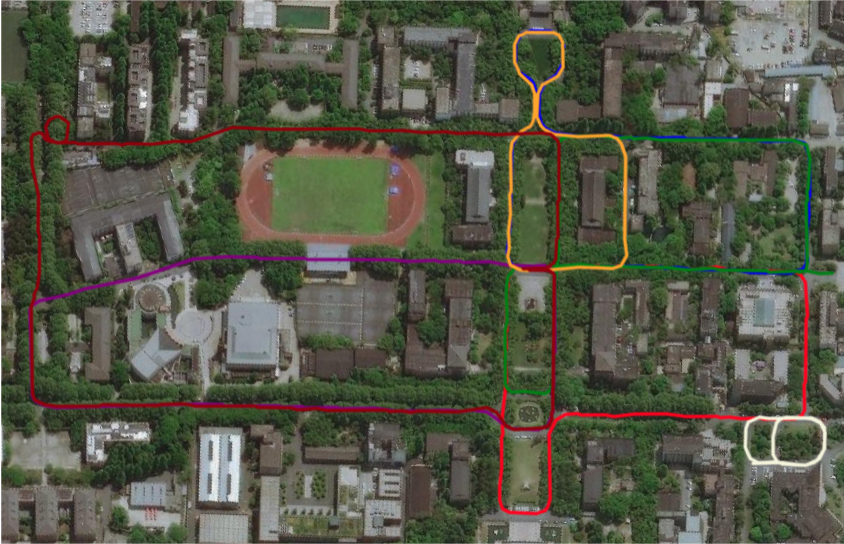}
    \captionsetup{font={small}}
	\caption{Seven trajectories of YQ dataset.}
	\label{fig:yq_traj}
\end{figure}

In the experiments, we evaluate the proposed continuous-time fixed-lag smoother in terms of pose estimation accuracy in various scenarios, systematically analyze the convergence speed and the accuracy of online timeoffset calibration, and disclose the runtime of the main stages in the proposed method. 
We assess a variety of sensor combinations powered by the continuous-time fixed-lag smoothing method, including
\begin{itemize}
    \item \textbf{CLIO}: one LIDAR and one IMU.
    \item \textbf{CLIO2}: two LiDARs and one IMU.
    \item \textbf{CLIC}: one LiDAR, one IMU and one camera.
    \item \textbf{CLIC2}: two LiDAR, one IMU and one camera.
\end{itemize}
We adopt the \textbf{A}bsolute \textbf{P}ose \textbf{E}rror (\textbf{APE}) as evaluation metric to compare the proposed method against three LI systems (CLINS~\cite{lv2021clins}, LIO-SAM~\cite{shan2020lio}), and three LIC systems (LVI-SAM~\cite{shan2021lvi}, VIRAL-SLAM~\cite{nguyen2021viral}, LIC-Fusion 2.0~\cite{zuo2020lic}) and two multi-LiADR systems (VIRAL-SLAM, MILIOM~\cite{nguyen2021miliom}).
In all experiments, the temporal knot distance $\Delta t$ is 0.03s, and the duration of LI temporal sliding window length is 0.12 seconds while the visual keyframe sliding window size is 10.

\subsection{Evaluation Datasets}
We evaluate the following three publicly-available datasets and our self-collected datasets:
\begin{itemize}
	\item \textbf{VIRAL}~\cite{nguyen2022ntu}:
	The Visual-Inertial-RAnging-Lidar Dataset contains a variety of sensors, including two 16-beam Ouster LiDARs at 10Hz, two monocular cameras at 10Hz, and a VectorNav VN100 IMU at 385Hz, etc. The dataset comprises nine sequences collected indoors and outdoors by a MAV (Micro Aerial Vehicle).
	\item \textbf{NCD}~\cite{ramezani2020newer}:
	The Newer College Dataset is collected using a handheld device that consists of a 64-beam Ouster LiDAR at 10Hz with its internal IMU at 100Hz, and a stereo camera at 30Hz. The dataset combines built environments, open spaces and vegetated areas.
	\item \textbf{LVI-SAM}~\cite{shan2021lvi}:
	The LVI-SAM Dataset features both handheld and vehicle (Jackal) platforms in outdoor open vegetated environments including geometrically degenerate surroundings with 16-beam LiDAR at 10Hz, camera at 20Hz, and IMU at 500Hz.
	\item \textbf{YQ}: The self-collected YuQuan Dataset collected on our university campus consists of seven sequences using an electric car with a sensor rig mounted on the top shown in Fig.~\ref{fig:lic_sensors}. The sensor rig comprises a 16-beam LiDAR at 10Hz, a camera at 20Hz, and an IMU at 400Hz. The trajectories of different sequences overlaid on Google map are shown in Fig.~\ref{fig:yq_traj}. GPS measurements are collected to provide groundtruth for this outdoor dataset.
    \item \textbf{Vicon Room}: The self-collected Vicon Room Dataset shares the identical sensor rig as YQ dataset. This indoor dataset is collected with hand-held random motion, and a motion capture system is leveraged to provide groundtruth trajectories.
\end{itemize}

\begin{table}[t]
    \centering
    \captionsetup{font={small}}
    \caption{The APE (RMSE, meter) results on YQ dataset (outdoor). The best result of LIC system without (or with) loop closure is \underline{underlined} (or in bold). }
    \resizebox{\linewidth}{!}{
    \begin{tabular}{@{}lccccc@{}}
    \toprule
    Sequence & \begin{tabular}[c]{@{}c@{}}LVI-SAM\\ (w/o loop)\end{tabular} & \begin{tabular}[c]{@{}c@{}}LIC-Fusion 2.0\\ (w/o loop)\end{tabular} & \begin{tabular}[c]{@{}c@{}}CLIC\\ (w/o loop)\end{tabular} & \begin{tabular}[c]{@{}c@{}}LVI-SAM\\ (w/ loop)\end{tabular} & \begin{tabular}[c]{@{}c@{}}CLIC\\ (w/ loop)\end{tabular} \\ \midrule
    YQ-01 (1005m) & \underline{1.614} & 3.300 & 1.826 & \textbf{1.227} & 1.537 \\
    YQ-02 (1021m) & 1.790 & 1.804 & \underline{1.626} & 1.610 & \textbf{1.363} \\
    YQ-03 (1058m) & 3.017 & 2.798 & \underline{2.616} & 1.886 & \textbf{1.701} \\
    YQ-04 (1233m) & 3.163 & 2.697 & \underline{2.450} & 2.402 & \textbf{1.917} \\
    YQ-05 (673m) & 1.721 & 1.550 & \underline{1.537} & 8.465 & \textbf{1.439} \\
    YQ-06 (1644m) & 3.753 & 3.862 & \underline{3.082} & 3.682 & \textbf{1.607} \\
    YQ-07 (414m) & 0.800 & 1.306 & \underline{0.761} & 0.795 & \textbf{0.761} \\  \bottomrule
    \end{tabular}
    }
    \label{tab:YQ_seq}
\end{table}

\begin{table}[t]
\centering
    \captionsetup{font={small}}
    \caption{The APE (RMSE, meter) results on VICON Room dataset (indoor). The best result is in bold. }
    \begin{tabular}{@{}cccc@{}}
    \toprule
    Seq. & \begin{tabular}[c]{@{}c@{}}LVI-SAM\\ (w/o loop)\end{tabular} & \begin{tabular}[c]{@{}c@{}}LIC-Fusion 2.0\\ (w/o loop)\end{tabular} & \begin{tabular}[c]{@{}c@{}}CLIC\\ (w/o loop)\end{tabular} \\ \midrule
     Seq1 (43m) & 0.393 & \textbf{0.033} & 0.080 \\
     Seq2 (84m) & 0.232 & 0.096 & \textbf{0.073} \\
     Seq3 (34m) &  0.364 & \textbf{0.052} & 0.073 \\
     Seq4 (53m) & 0.395 & 0.092 & \textbf{0.089} \\
     Seq5 (50m) & 0.155 & \textbf{0.044} & 0.171 \\
     Seq6 (88m) & 0.459 & \textbf{0.046} & 0.128 \\ \bottomrule
    \end{tabular}
    \label{tab:viconRoom}
\end{table}

\begin{figure}[t]
    \centering
    \includegraphics[width=1\linewidth]{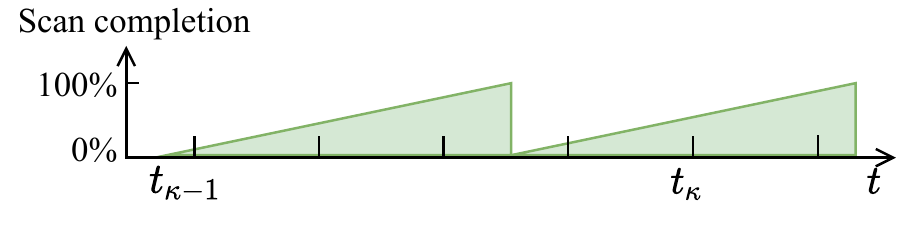}
    \vspace{-1em}
    \captionsetup{font={small}}
    \caption{
        A LiDAR-inertial temporal sliding window over $\left[t_{\kappa-1}, t_\kappa\right)$ consists of multiple LiDAR scans (green blocks), which can be an incomplete scan. }
    \label{fig:clic_scan_percent}
\end{figure}

\subsection{LiDAR-Inertial Fusion: CLIO}
In this experiment, we evaluate the accuracy of continuous-time trajectory estimation of the CLIO system enabled by the proposed continuous-time fixed-lag smoothing. The results on VIRAL dataset are shown in Tab.~\ref{tab:viral_ape}, CLIO outperforms other LI methods in most sequences and achieves an average RMSE of 0.034m. CLINS is much more accurate than all the discrete-time methods, including LIO-SAM~\cite{shan2020lio}, MILIOM~\cite{nguyen2021miliom}, and VIRAL~\cite{nguyen2021viral}.
Compared to continuous-time method, CLINS~\cite{lv2021clins}, which only optimizes control points of trajectory within the time span of the newest LiDAR scan and lacks probabilistic marginalization, the proposed CLIO, optimizing all the control points in a sliding window involved with several LiDAR scans (see Fig.~\ref{fig:timeline} and Fig.~\ref{fig:clic_scan_percent}) and benefiting from marginalization, can achieve higher accuracy.

We further conduct evaluations in large-scale scenarios on the LVI-SAM dataset and NCD dataset (Tab.~\ref{tab:lvi_ape} and \ref{tab:ncd_ape}), CLIO achieves higher accuracy on most sequences compared to LIO-SAM.
Especially, it is worth noting that on the NCD\_06 sequence with the handheld sensor shaking vigorously, CLIO shows significant superiority over the discrete-time LIO-SAM.
The improvement in accuracy could be attributed to our continuous-time trajectory formulation as it adequately addresses the motion distortion of LiDAR point cloud, which is of significant importance in highly-dynamic motion scenarios.

The presented continuous-time fixed-lag smoothing method supports fusion of multiple sensors conveniently. In Tab~\ref{tab:viral_ape}, We also showcase the pose estimation accuracy of a system with the fusion of IMU and two LiDARs, dubbed CLIO2. We can see that CLIO2 has on-par accuracy with CLIO on most of the sequences, and shows more accurate pose estimation over the outdoor sequences (nya\_xx).

\begin{figure}[t]
	\centering
    \includegraphics[width=1\linewidth]{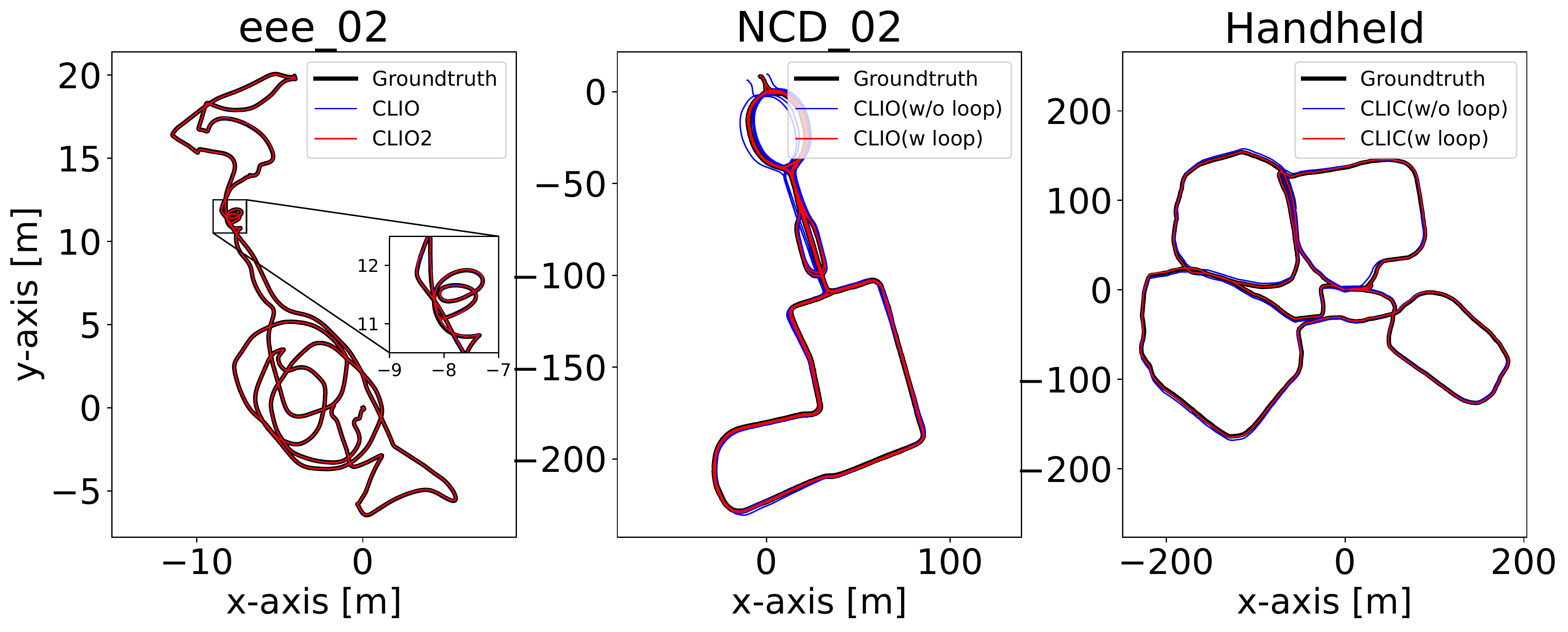}
    \captionsetup{font={small}}
	\caption{The estimated trajectories compared to the groundtruth in eee\_02 sequence of VIRAL dataset, NCD\_02 sequence of NCD dataset and Handheld sequence of LVI-SAM dataset.}
	\label{fig:lvi_traj}
\end{figure}

\begin{figure}[t]
	\centering
    \includegraphics[width=1\linewidth]{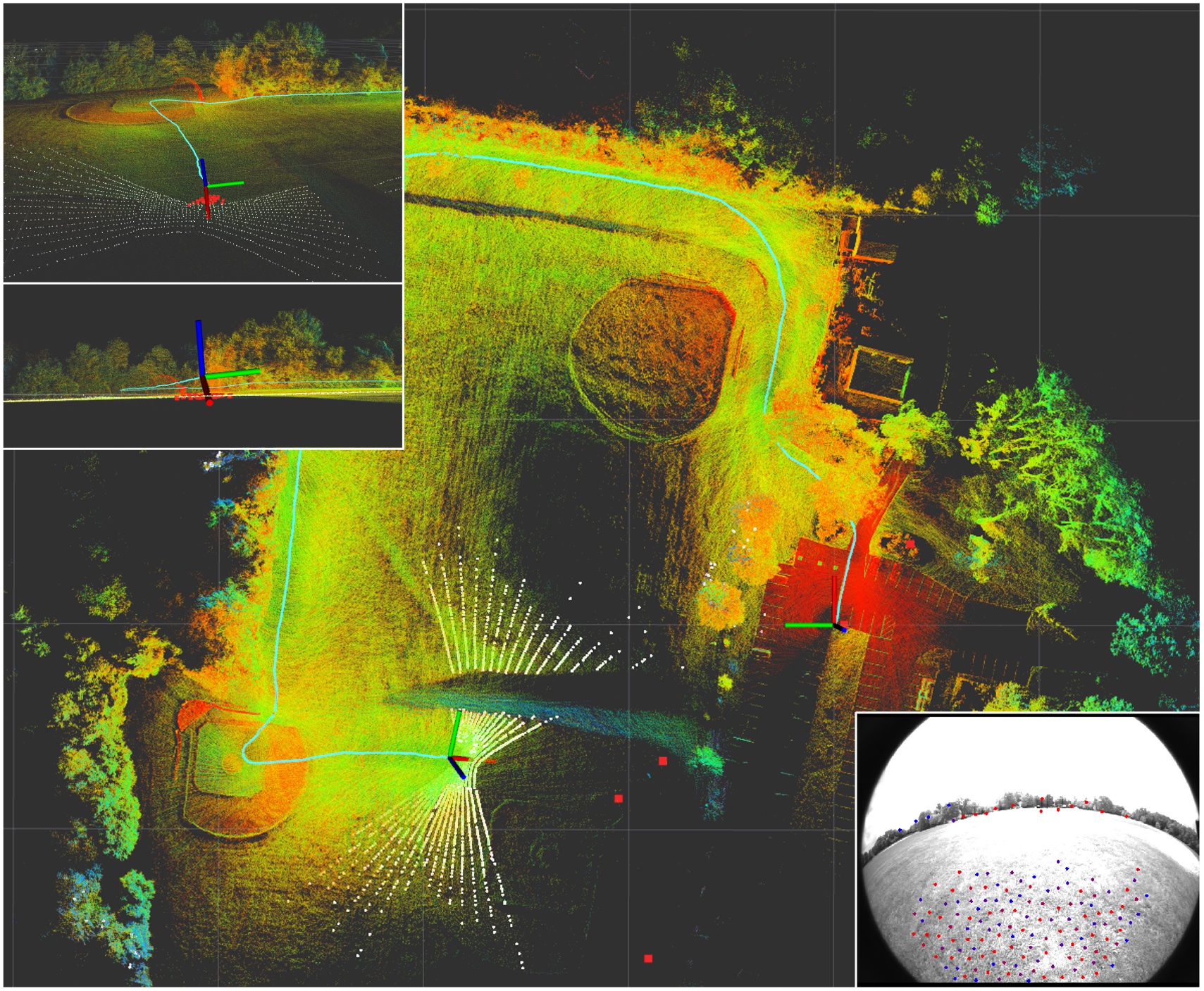}
    \captionsetup{font={small}}
	\caption{ The LiDAR map (color points) and visual landmarks (red squares) during the system passing open areas when running Handheld sequence. The current scan (white points) only observes the ground while camera can track stable visual features.}
	\label{fig:lvi_map}
	\vspace{-1em}
\end{figure}

\begin{figure}[t]
	\centering
    \includegraphics[width=1\linewidth]{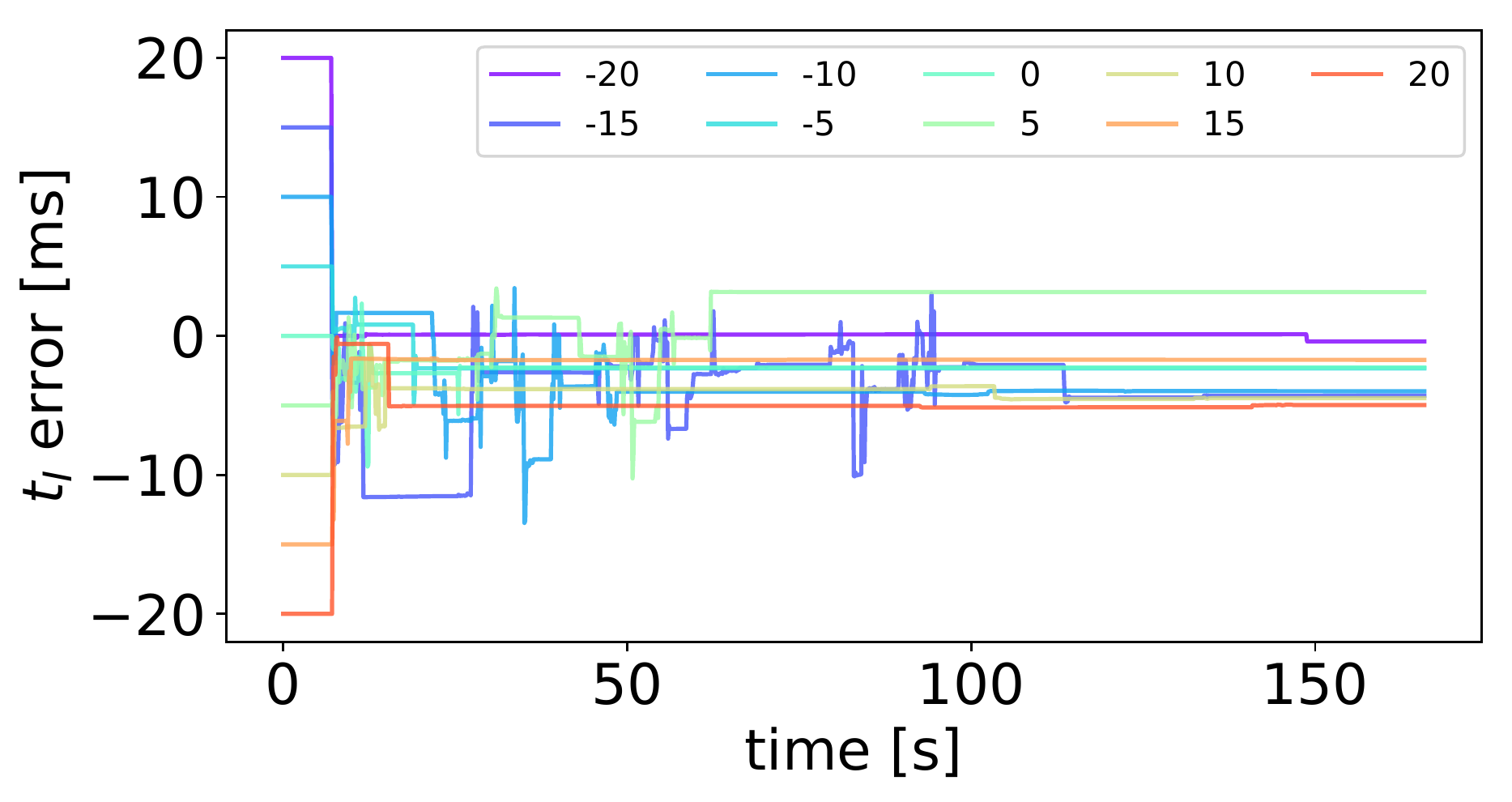}
    \captionsetup{font={small}}
	\caption{Temporal calibration error of CLIC system in NCD\_01 sequence of NCD dataset. Although we start from various initial values of timeoffsets, the estimates of timeoffsets are able to quickly converge.}
	\label{fig:NCD_calib}
\end{figure}

\subsection{LiDAR-Inertial-Camera Fusion: CLIC}
In this section, we discuss the accuracy of the LiDAR-Inertial-Camera pose estimation system enabled by continuous-time fixed-lag smoothing.
The experimental results on the VIRAL and LVI datasets are summarized in Tab.~\ref{tab:viral_ape} and Tab.~\ref{tab:lvi_ape}. In addition, the evaluation results on our self-collected indoor and outdoor datasets are shown in Tab.~\ref{tab:YQ_seq} and~\ref{tab:viconRoom}. 
CLIC tightly fuses LiDAR, IMU and camera measurements, and successfully generates good pose estimation in Handheld sequences collected over large open areas while CLIO fails, achieving significantly higher accuracy compared to the discrete-time method LVI-SAM~\cite{shan2021lvi}. It indicates that visual factors fused into CLIO can help improve the system's applicability. Figure.~\ref{fig:lvi_map} showcases a snapshot on the Handheld sequence where the LiDAR could only detect the ground while camera can track stable visual features to further constrain the optimization problem. 
As discussed in Sec.~\ref{sec:keyframe_sliding_win}, we only optimize the control points within the LI temporal sliding window for computation efficiency reasons, and the other control points involved in visual keyframe sliding window are kept fixed during optimization, which might degrade the effects of visual factors.
In Tab.~\ref{tab:YQ_seq} and ~\ref{tab:viconRoom}, LIC-Fusion 2.0~\cite{zuo2020lic} shows pleasing performance in both indoor and outdoor scenarios, which benefits from reliable sliding-window plane-feature tracking. However, we can see that LIC-Fusion 2.0 has degraded performance in outdoor scenarios (see Tab.~\ref{tab:YQ_seq}). The possible reason is that LIC-Fusion 2.0 relies on stably-tracked plane features to update LiDAR poses, while the outdoor scenarios are filled with tree clumps, and can fail to provide sufficient structural planes compared to indoors.

The accuracy of CLIC can be further improved when enabling online timeoffset calibration between different sensors (see Tab.~\ref{tab:lvi_ape}). In Fig.~\ref{fig:lvi_traj}, we also show some representative estimated trajectories of different methods aligned with the ground truth trajectories.

\begin{table}[t]
    \centering
    \captionsetup{font={small}}
	\caption{Timing of different modules of CLINS, CLIO and CLIC in eee\_01 sequence with a duration of 397 seconds. }
	\label{tab:eee01_timing}
    \begin{tabular}{@{}cccc@{}}
    \toprule
     & CLINS & CLIO & CLIC \\ \midrule
    Update Local Map & 17.39 & 11.56 & 11.52 \\
    Update Trajectory & 1184.34 & 58.55 & 80.64 \\
    Update Prior & 0.00 & 8.45 & 8.44 \\
    Others & 400.13 & 139.27 & 193.97 \\ \midrule
    Total Time Cost (second) & 1601.86 & 217.82 & 294.57 \\ \bottomrule
    \end{tabular}
\end{table}

\subsection{Online Temporal Calibration}
In this section, we examine the performance of online temporal calibration in the proposed multi-sensor fusion systems. Experiments are conducted on the NCD\_01 sequence of NCD dataset~\cite{ramezani2020newer}. 
We manually add additional timeoffsets of $-20\sim 20$ ms to the raw timestamps, and the timeoffest $t_I$ provided from NCD dataset is 0ms.
Fig.~\ref{fig:NCD_calib} shows the error of estimated timeoffset over time. We start to estimate the timeoffset 5 seconds after the start of the system, and the IMU timeoffset converges quickly, with most of the trials converging within 3 seconds. In addition, the final estimated timeoffset mean(std) is -2.0ms(2.7ms).

\subsection{Runtime analysis}
We investigate runtime of the main modules in CLIO and CLIC in eee\_01 sequence of VIRAL dataset~\cite{nguyen2021viral}. Importantly, we notice the average runtime of proposed method remains almost constant whether in a long sequence or a short sequence. The method is implemented in C++ and executed on the desktop PC with an Intel i7-7700K and 32GB RAM, and Tab.~\ref{tab:eee01_timing} summarizes the time consumption of CLINS~\cite{lv2021clins}, CLIO and CLIC. The term \textit{Update Local Map} in Tab.~\ref{tab:eee01_timing} represents update local LiDAR map for associating LiDAR feature, \textit{Update Trajectory} represents solving the problem in Eq.~\ref{eq:least-squares} and \textit{Update Trajectory} represents the marginalization process. The term \textit{Others} includes feature extraction of image and LiDAR cloud, trajectory initialization, data association, et al.
With analytical Jacobian computations for optimization and using sliding window and marginalization to bound computation complexity by the continuous-time fixed-lag smoothing, the enabled multi-sensor fusion methods (CLIO and CLIC) achieve real-time capability. In contrast, the continuous-time method, CLINS~\cite{lv2021clins}, consumes much more time and fails to run in real time. Here, real-time performance refers to the total elapsed time to process measurements from the sensors is less than the sensor data collection time.
We also note that our current implementation is not optimal, and there is still significant space to further improve efficiency. 

\section{Conclusion}
\label{sec:conclusion}

This paper exploits continuous-time fixed-lag smoothing for asynchronous multi-sensor fusion in a factor graph framework.
Specifically, we propose to probabilistically marginalize old states and measurements out of the sliding window, and derive analytic Jacobians for continuous-time optimization. 
Benefiting from the nature of continuous-time trajectory formulation, heterogeneous multi-sensor measurements at any time instants can be seamlessly fused.
Empowered by the continuous-time fixed-lag smoothing, we design the estimators at different sensor configurations, for tight fusion of multiple LiDARs, IMU and camera sensors. Our estimators, including LiDAR-Inertial systems and LiDAR-Inertial-Camera systems, show significant advances in pose estimation accuracy over the existing state-of-the-art methods. Online temporal calibration between sensors is also naturally supported in the continuous-time estimator. 
We demonstrate the accuracy and applicability of our method on three available public datasets and compare it with the state-of-art LiDAR-Inertial, LiDAR-Inertial-Camera and multi-LiDAR systems, respectively. 
Future work includes more efficient management of LiDAR map~\cite{bai2022faster}, utilizing more reliable LiDAR feature tracking and estimation method~\cite{zuo2020lic}, and exploring the potential benefits of non-uniform B-spline.

\section{Acknowledgement}
This work is supported by NSFC 62088101 Autonomous Intelligent Unmanned Systems. We would like to thank Kewei Hu and Xiangrui Zhao for fruitful discussion.

{
\vspace{0.05cm}
\def\bibfont{\scriptsize}
\printbibliography
}

\newpage
\section{Appendix} \label{sec:appendix}

In the following, we first derive the Jacobians of the continuous-time trajectory value w.r.t control points (Sec.~\xing{IV-A}), then derive the Jacobians of residuals w.r.t trajectory value (Sec.~\xing{IV-B, IV-C, IV-D, IV-E}) and finally use chain rule to get the Jacobians of residuals w.r.t estimated state (Eq.~(\xing{11})). All the analytic Jacobians are utilized in the factor graph optimization (Eq.~(\xing{12})), which significantly speeds up the problem-solving.  
The number of control points in the error state is
\begin{align}
    N_\phi = \eta+k-1
\end{align}
where $k$ is B-Spline order and $\eta$ is the number of knot intervals in the temporal sliding window (see Sec.~\xing{V-A2}). Note that, not every control point in the state vector (Eq.~(\xing{11})) is involved in a specific factor, thus we define there are $N_\phi^{\text{pre}}$ (or $N_\phi^{\text{post}}$) control points before (or after) the involved control points. Here, we also define $N_\ell$ as the number of landmarks in the state.

\subsection{Jacobian for Control Points}
\label{sec:appendix_ctrl_points}

We follow the notation in~\cite{sommer2020efficient} to give the Jacobians of the trajectory value to the control points: 
\begin{equation}
    \frac{\partial \mathbf{f}}{\partial \bR_{i+j}} = 
    \ddj{\mathbf{f}}  \cdot \frac{\partial{\bd_j}}{\partial \bR_{i+j}} + \frac{\partial \mathbf{f}}{\partial\bd_{j+1}} \cdot \frac{\partial{\bd_{j+1}}}{\partial \bR_{i+j}}\,
\end{equation}
for $j>0$ and for $j=0$ we obtain
\begin{equation}
    \frac{\partial \mathbf{f}}{\dif \bR_{i}} = \frac{\partial \mathbf{f}}{\partial\bR_i} + \frac{\partial \mathbf{f}}{\partial\bd_{1}} \cdot \frac{\partial{\bd_1}}{\partial \bR_{i}}\,,
\end{equation}
where $\mathbf{f}$ is the queried pose or velocity. Note that we follow the recursive scheme in~\cite{sommer2020efficient} to efficient compute Jacobians of queried pose $\{\frac{\partial {}^G_I\bR(\tau_{\ell})} {\partial \tilde{\bx}_R^{\kappa}}, \frac{\partial  {}^G\bp_I(\tau_{\ell}) }{\partial \tilde{\bx}_p^{\kappa}}\}$, 
angular velocity $\frac{\partial {}^I\bomega(\tau_m)} {\partial \tilde{\bx}_R^{\kappa}}$ 
and linear acceleration $\frac{\partial {}^G\ba(\tau_m)} {\partial \tilde{\bx}_R^{\kappa}}$.

\subsection{Jacobians for IMU Factor}
\label{sec:appendix_imu_factor}
Given the IMU factor defined at Eq.~(\xing{18}), the Jacobians w.r.t error state is 
\begin{equation}
\begin{aligned}
    \bH_{I_m}^{\kappa} &= \frac{\partial \br_{I}} {\partial \tilde{\bx}}  \\ 
    &= 
    \begin{bmatrix}
    \bH_{I_m}^{\bx_R} & 
    \bH_{I_m}^{\bx_p} & 
    \bH_{I_m}^{\bx_{I_b}} & 
    \bzero_{6\times N_l} & 
    \bH_{I_m}^{t_I} &
    \bzero_{6\times 1}
    \end{bmatrix}
\end{aligned}
\end{equation}
where
\begin{align}
    \bH_{I_m}^{\bx_R} 
    &= \begin{bmatrix}
        \bzero_{3\times N_\phi^{\text{pre}}} & \frac{\partial {}^I\bomega(\tau_m)} {\partial \tilde{\bPhi}_R(\tau_m)} & \bzero_{3\times N_\phi^{\text{post}}} \\
        \bzero_{3\times N_\phi^{\text{pre}}} & \textcolor{black}{\Pi_{i1}}
        \frac{\partial {}^G\ba(\tau_m)} {\partial \tilde{\bPhi}_R(\tau_m)} & \bzero_{3\times N_\phi^{\text{post}}}
    \end{bmatrix} \,,\\
    \bH_{I_m}^{\bx_p} 
    &= \begin{bmatrix}
        \bzero_{3\times N_\phi^{\text{pre}}} & \bzero_{3\times k} & \bzero_{3\times N_\phi^{\text{pre}}}\\
        \bzero_{3\times N_\phi^{\text{pre}}} & \textcolor{black}{\Pi_{i2}} 
        \frac{\partial {}^I\ba(\tau_m)} {\partial \tilde{\bPhi}_p(\tau_m)} 
        & \bzero_{3\times N_\phi^{\text{post}}}
    \end{bmatrix} \,,\\
    \bH_{I_m}^{\bx_{I_b}} 
    &= \begin{bmatrix}
    \bzero_{3\times 3} & \bzero_{3\times 3} & \bI_{3\times 3} & \bzero_{3\times 3} \\
    \bzero_{3\times 3} & \bzero_{3\times 3} &  \bzero_{3\times 3} & \bI_{3\times 3}
    \end{bmatrix} \,,\\
    \bH_{I_m}^{t_I} 
    &= \begin{bmatrix}
    {}^I\dot{\bomega}(\tau_m) \\
    {}^I\dot{\ba}(\tau_m)
    \end{bmatrix} \,,
\end{align}
and 
\begin{align}
    \textcolor{black}{\Pi_{i1}} &= \lfloor ^G_I\bR^{\top}(\tau_m){}^G\ba(\tau_m) \times \rfloor \\
    \textcolor{black}{\Pi_{i2}} &= {}^G_I\bR^{\top}(\tau_m) \,.
\end{align}

\subsection{Jacobians for Bias Factor}
\label{sec:appendix_bias_factor}
Given the Bias factor defined at Eq.~(\xing{19}), the Jacobians w.r.t error state is
\begin{equation}
\begin{aligned}
    \bH_{I_b}^{\kappa} &= \frac{\partial \br_{I}} {\partial \tilde{\bx}}\\
    &= 
    \begin{bmatrix}
    \bzero_{6\times N_\phi} & 
    \bzero_{6\times N_\phi} & 
    \bH_{I_b}^{\bx_{I_b}} & 
    \bzero_{6\times N_l} & 
    \bzero_{6\times 2}
    \end{bmatrix}
\end{aligned}
\end{equation}
where
\begin{align}
    \bH_{I_b}^{\bx_{I_b}} = \begin{bmatrix}
    -\bI_{3\times 3} & \bzero_{3\times 3} & \bI_{3\times 3} & \bzero_{3\times 3} \\
    \bzero_{3\times 3} & -\bI_{3\times 3} &  \bzero_{3\times 3} & \bI_{3\times 3}
    \end{bmatrix} \,.
\end{align}

\subsection{Jacobians for LiDAR Factor}
\label{sec:appendix_lidar_factor}
Given the LiDAR factor defined at Eq.~(\xing{14}), the Jacobians w.r.t error state is
\begin{equation}
\begin{aligned}
    \bH_{\ell} &= \frac{\partial \br_{L}}{\partial{}^{G}\hat{\bp}_{\ell} }\,
    \frac{\partial{}^{G}\hat{\bp}_{\ell} }{\partial \tilde{\bx}} \\
    &= {}^{G}\bn^\top_{\pi} 
    \begin{bmatrix}
    \bH_{\ell}^{\bx_R} & 
    \bH_{\ell}^{\bx_p} & 
    \bzero_{3\times 12} & 
    \bzero_{3\times N_l} & 
    \bzero_{3\times 2}
    \end{bmatrix}
\end{aligned}
\end{equation}
where
\begin{align}
    \bH_{\ell}^{\bx_R} &= 
    \begin{bmatrix} 
    \bzero_{3\times N_\phi^{\text{pre}}} &
    \textcolor{black}{\Pi_{\ell 1}}
    \,
    \frac{\partial {}^G_L\bR(\tau_{\ell})} {\partial \tilde{\bPhi}_R(\tau_\ell)} 
    & \bzero_{3\times N_\phi^{\text{post}}} 
    \end{bmatrix} \,, \\
    \bH_{\ell}^{\bx_p} &=
    \begin{bmatrix} 
    \bzero_{3\times N_\phi^{\text{pre}}} &
    \frac{\partial  {}^G\bp_L(\tau_{\ell}) }{\partial \tilde{\bPhi}_p(\tau_\ell)} 
    & \bzero_{3\times N_\phi^{\text{post}}} 
    \end{bmatrix} \,,
\end{align}
and 
\begin{align}
    \textcolor{black}{\Pi_{\ell 1}} &= -{}^G_L\bR(\tau_{\ell}) \lfloor {}^L\bp_{\ell}  \times \rfloor \,.
\end{align}

\subsection{Jacobians for Visual Factor}
\label{sec:appendix_vidual_factor}

The visual factor (Eq.~(\xing{21})) is related to two poses at time instants, $\tau_a$ and $\tau_b$, and the Jacobians w.r.t error state is
\begin{equation}
\begin{aligned}  
    \bH_{\kappa}^{s_n} = 
    \frac{\partial \br_{c}}{\partial \hat{\bp}_{{\kappa}}^{s_n}}
    \begin{bmatrix}
    \bH_{\kappa}^{\bx_R} & 
    \bH_{\kappa}^{\bx_p} & 
    \bzero_{3\times 12} & 
    \bH_{\kappa}^{\bx_{\lambda}} & 
    \bzero_{3\times 1} &
    \bH_{\kappa}^{t_C}
    \end{bmatrix}
\end{aligned}
\end{equation}
where
\begin{align}
    \frac{\partial \br_{c}}{\partial \hat{\bp}_{{\kappa}}^{s_n}} &= \frac{1}{\be_{3}^{\top} \hat{\bp}_{{\kappa}}^{s_n}} \begin{bmatrix}
    1 & 0 & -\left({\be_{1}^{\top} \hat{\bp}_{{\kappa}}^{s_n}}\right)/\left({\be_{3}^{\top} \hat{\bp}_{{\kappa}}^{s_n}}\right) \\
    0 & 1 & -\left({\be_{2}^{\top} \hat{\bp}_{{\kappa}}^{s_n}}\right)/\left({\be_{3}^{\top} \hat{\bp}_{{\kappa}}^{s_n}}\right)
    \end{bmatrix} \,,\\
    \bH_{\kappa}^{\bx_R} &=
    \begin{bmatrix} 
    \bzero_{3\times N_\phi^{\text{pre}}} &
     \lfloor \textcolor{black}{\Pi_{c1}} \times \rfloor \, \frac{\partial {}^G_C\bR(\tau_b)} {\partial \tilde{\Phi}_R(\tau_b)} 
    & \bzero_{3\times N_\phi^{\text{post}}} 
    \end{bmatrix} \notag \\
    & + \begin{bmatrix} 
    \bzero_{3\times N_\phi^{\text{pre}}} & \textcolor{black}{\Pi_{c2}} \, \frac{\partial {}^G_C\bR(\tau_a)} {\partial \tilde{\Phi}_R(\tau_a)} 
    & \bzero_{3\times N_\phi^{\text{post}}} 
    \end{bmatrix}  \,, \\
    \bH_{\kappa}^{\bx_p} &=    
    \begin{bmatrix} 
    \bzero_{3\times N_\phi^{\text{pre}}} &
     \textcolor{black}{\Pi_{c3}} \, \frac{\partial {}^G\bp_C(\tau_b)} {\partial \tilde{\Phi}_p(\tau_b)} 
    & \bzero_{3\times N_\phi^{\text{post}}} 
    \end{bmatrix} \notag \\
    & + \begin{bmatrix} 
    \bzero_{3\times N_\phi^{\text{pre}}} & \textcolor{black}{\Pi_{c4}} \, \frac{\partial {}^G\bp_C(\tau_a)} {\partial \tilde{\Phi}_p(\tau_a)} 
    & \bzero_{3\times N_\phi^{\text{post}}} 
    \end{bmatrix}  \,, \\
    \bH_{\kappa}^{\bx_{\lambda}} & =\frac{-1}{\lambda^2_{{\kappa}}} \cdot
    {}^{C_b}_{C_a}\bR \cdot \pi_c( \brho_{\kappa}^{a} ) \,, \\
    \bH_{\kappa}^{t_C} &= 
    {}^G_C\bR(\tau_{b})^{\top}\Big({}^G\bv_I(\tau_a)-{}^G\bv_I(\tau_b)\Big) \notag \\
    &+ {}^G_C\dotbR(\tau_b)^{\top}\Big({}^G_C\bT(\tau_{a})\cdot \frac{1}{\lambda_{\jmath}} \pi_c( \brho_{\jmath}^{a}) \Big) \notag \\
    &+ {}^G_C\bR(\tau_b)^{\top}\Big({}^G_C\dotbR(\tau_{a})\cdot \frac{1}{\lambda_{\jmath}} \pi_c( \brho_{\jmath}^{a}) \Big) 
\end{align}
and 
\begin{align}
    \textcolor{black}{\Pi_{c1}} &= -{}^G_C\bR(\tau_{b})^{\top} \Big({}^G_C\bT(\tau_{a})\cdot \frac{1}{\lambda_{\jmath}} \pi_c( \brho_{\jmath}^{a}) - {}^G\bp_C(\tau_{b})\Big) \\
    \textcolor{black}{\Pi_{c2}} &= -{}^G_C\bR(\tau_{b})^{\top} \Big({}^G_C\bR(\tau_{a}) \cdot  \lfloor \frac{1}{\lambda_{\jmath}} \pi_c( \brho_{\jmath}^{a}) \times \rfloor \Big) \\
    \textcolor{black}{\Pi_{c3}} &= -{}^G_C\bR(\tau_{b})^{\top} \\
    \textcolor{black}{\Pi_{c4}} &= {}^G_C\bR(\tau_{b})^{\top}{}^G_C\bR(\tau_{a}) \,.
\end{align}

\end{document}